\documentclass[runningheads]{llncs}

\usepackage{eccv}

\usepackage{eccvabbrv}

\usepackage{graphicx}
\usepackage{booktabs}

\usepackage{multirow} 
\usepackage{booktabs} 

\usepackage[table]{xcolor}

\usepackage[accsupp]{axessibility}  

\usepackage{hyperref}
\usepackage{orcidlink}
\usepackage{marvosym}

\makeatletter
\DeclareRobustCommand{\corrauthmark}{\Letter}
\renewcommand{\@fnsymbol}[1]{%
  \ifcase#1
  \or \textdagger
  \or \corrauthmark
  \or \textdaggerdbl
  \or \S
  \or \P
  \else \@ctrerr
  \fi}
\makeatother

\begin{document}

\title{DecepGPT: Schema-Driven Deception Detection with Multicultural Datasets and Robust Multimodal Learning} 

\titlerunning{DecepGPT}


\author{
Jiajian Huang\inst{1}\thanks{Equal contribution.}
\and Dongliang Zhu\inst{2}\protect\footnotemark[1]
\and Zitong Yu\inst{1,4}\thanks{Corresponding authors.}
\and Hui Ma\inst{1}\protect\footnotemark[2]
\and Jiayu Zhang\inst{1}
\and Chunmei Zhu\inst{3}
\and Xiaochun Cao\inst{3}
}

\authorrunning{J.~Huang et al.}

\institute{
Great Bay University, China\\
\email{\{yuzitong,mah\}@gbu.edu.cn}
\and
Wuhan University, China
\and
Sun Yat-sen University, China
\and
Dongguan Key Laboratory for Intelligence and Information Technology, China
}

\maketitle

\begin{abstract}
Multimodal deception detection aims to identify deceptive behavior by analyzing audiovisual cues for forensics and security. In these high-stakes settings, investigators need verifiable evidence connecting audiovisual cues to final decisions, along with reliable generalization across domains and cultural contexts. However, existing benchmarks provide only binary labels without intermediate reasoning cues. Datasets are also small with limited scenario coverage, leading to shortcut learning. We address these issues through three contributions. First, we construct reasoning datasets by augmenting existing benchmarks with structured cue-level descriptions and reasoning chains, enabling models to output auditable reports. Second, we release T4-Deception, a multicultural dataset based on the unified ``To Tell the Truth'' television format implemented across four countries. With 1695 samples, it is the largest non-laboratory deception detection dataset. Third, we propose two modules for robust learning under small-data conditions. Stabilized Individuality-Commonality Synergy (SICS) refines multimodal representations by combining learnable global priors with sample-adaptive residuals and applying polarity-aware recalibration. Distilled Modality Consistency (DMC) aligns modality-specific predictions with the fused multimodal predictions via knowledge distillation to prevent unimodal shortcut learning. Experiments on three established benchmarks and our novel dataset demonstrate that our method achieves state-of-the-art performance in both in-domain and cross-domain scenarios, while exhibiting superior transferability across diverse cultural contexts. The datasets and code are available at \href{https://github.com/open-code-and-source/DecepGPT}{this link}.
\end{abstract}

\section{Introduction}
Multimodal deception detection (MDD) aims to identify deceptive behavior by analyzing audio and visual cues\cite{perez2015deception, wu2018deception,zhu2026svc}, which provide objective decision support in high-stakes social analysis, such as forensic investigation and security screening\cite{guo2023dolos}, where human judgment is often subject to cognitive bias\cite{bond2006accuracy}.

\begin{figure}[t]
    \centering
    \includegraphics[width=\columnwidth]{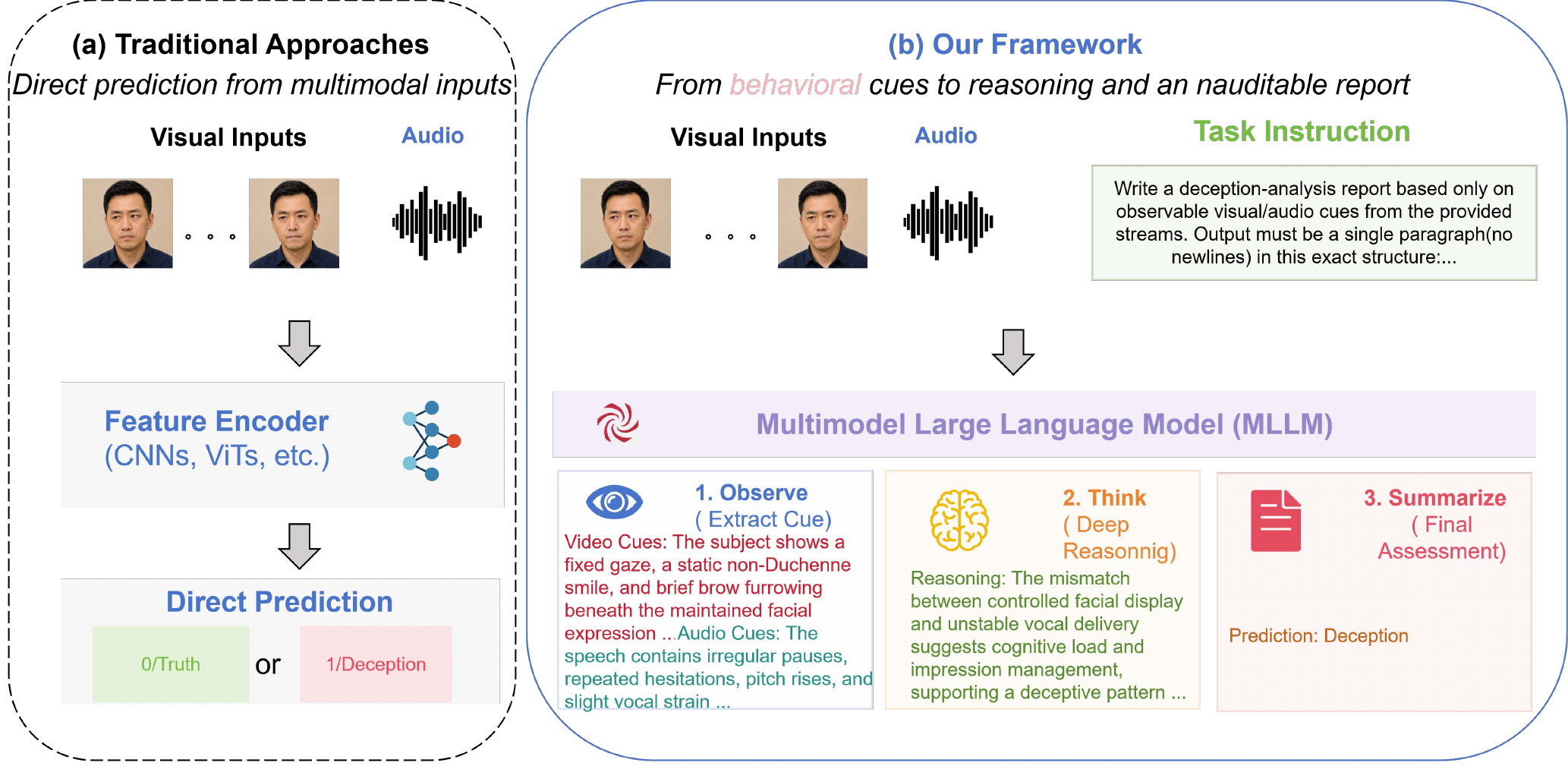}
    \caption{Comparison between the traditional paradigm and our auditable framework.
    (a) Traditional methods directly map behavioral signals to binary labels, leaving the decision process opaque.
    (b) Our method generates structured reports with audiovisual cues and reasoning paths, making the final judgment auditable.}
    \label{compared_method}
\end{figure}

Recent progress in MDD has evolved from handcrafted behavioral descriptors\cite{perezrosas2015verbal} to end-to-end audiovisual deep learning models \cite{nam2023facialcuenet,zhang2026multimodal}. However, as illustrated in Fig.~\ref{compared_method}a, traditional MDD methods are predominantly label-centric, focusing on optimizing binary classification accuracy. While achieving competitive performance, these methods typically provide only a final binary prediction. In forensic and legal contexts, a standalone label is insufficient. Investigators need to understand why a sample is flagged as deceptive, with evidence connecting behavioral cues such as micro-expressions and voice prosody to the final decision. Furthermore, MDD methods must demonstrate generalization across diverse cultural contexts to satisfy the requirements of real-world applications\cite{taylor2014cross}.

Existing benchmarks\cite{guo2023dolos,gupta2019bagoflies} provide only binary labels without intermediate reasoning cues, preventing models from producing verifiable evidence. Moreover, existing datasets have limited scenario coverage. Important factors, such as identity pretense and cross-cultural variations, are still not well studied, which limits the generalization ability of MDD methods. In addition, the small scale of available data often causes models to learn spurious correlations\cite{geirhos2020shortcut} during training.

We address these issues through three contributions. First, we construct a reasoning dataset by augmenting existing benchmarks with structured cue-level descriptions and reasoning chains, enabling the generation of auditable reports as shown in Fig.~\ref{compared_method}b. Second, we release T4-Deception, a multicultural dataset covering identity pretense across four countries (the U.S., Germany, Vietnam, and Bulgaria) under a unified 'To Tell the Truth' television format. With 1695 samples, it is the largest non-laboratory deception benchmark to date. Third, we propose two modules for robust learning under small-data conditions. Stabilized Individuality-Commonality Synergy (SICS) refines multimodal features through a polarity-aware adjustment mechanism, which synergizes a learnable global prior with sample-adaptive residuals to enhance or suppress specific feature dimensions. Meanwhile, Distilled Modality Consistency (DMC) introduces a consistency regularizer that aligns unimodal predictions with multimodal teacher predictions via knowledge distillation to mitigate shortcut learning.

In summary, the main contributions of this article are as follows:
\begin{itemize}
\item \textbf{Reasoning Dataset.} We provide a standardized pipeline to enrich existing benchmarks with structured audiovisual cues and reasoning chains, enabling the generation of auditable reports for verifiable decision-making.

\item \textbf{Multicultural Dataset.} We release T4-Deception (To Tell the Truth across 4 cultures), a large-scale dataset covering identity pretense across four countries under a unified television format. With 1695 samples, it is currently the largest non-laboratory dataset in the field.

\item \textbf{Robust Multimodal Learning Modules.} We propose Stabilized Individuality-Commonality Synergy (SICS) for polarity-aware feature refinement and Distilled Modality Consistency (DMC) for modality consistency distillation, which effectively improve both in-domain and cross-domain performance.

\end{itemize}

\section{Related Work}
\label{sec:related}
\textbf{Multimodal Deception Detection.} 
Early MDD studies relied on handcrafted behavioral cues, including facial expressions, gestures, prosody, and linguistic indicators~\cite{perezrosas2015verbal}. Later methods adopted deep audiovisual models to capture fine-grained deception patterns, using CNN/ResNet-based visual encoders~\cite{wu2018deception,karnati2022lienet,ding2019face} and attention-, graph-, or video-based multimodal architectures~\cite{csen2020multimodal,hsiao2022attention,zhang2022fine,zhuo2024video}. Recent works further improve cross-modal learning~\cite{guo2023dolos} and generalization~\cite{guo2024benchmarking, Lin2026FromUT}.  AFAFFAKT~\cite{Ji2025AFFAKTAH} explores affective knowledge transfer to mitigate the small-data challenge in deception detection. However, most existing methods still formulate MDD as binary truthful/deceptive classification, offering limited evidence-level reasoning or auditability. Although MLLMs provide new opportunities for semantic reasoning and interpretable decision support~\cite{liu2023llava}, they remain vulnerable to hallucinated rationales in deception scenarios~\cite{miah2025hidden}. We address this issue by augmenting deception detection datasets with structured reasoning chains and training model to standardize evidence extraction for audition.

\noindent\textbf{MLLM Reasoning.}
Recent MLLMs can generate free-form chain-of-thought (CoT) or rationale-style explanations for visual and video reasoning tasks~\cite{zhang2023multimodal,maaz2024videochatgpt}. However, in small-scale multimodal deception detection, such reasoning may be vulnerable to overfitting, unimodal shortcuts, and hallucinated rationales, especially when subtle behavioral cues such as facial dynamics, micro-gestures~\cite{wang2026micro}, and vocal patterns are weak or ambiguous. To address this, we formulate MLLM reasoning as schema-constrained audiovisual evidence tracing, and introduce SICS to stabilize multimodal representations and DMC to reduce unimodal shortcut learning.

\noindent\textbf{Behavioral Decoupling and Stabilized Refinement.}
Identity- or sample-specific variations may obscure shared behavioral cues~\cite{hazarika2020misa}. Conventional decomposition~\cite{hu2021learning} and modulation paradigms~\cite{arevalo2017gated} provide ways to separate features or adaptively weight samples. Motivated by this, SICS focuses on stabilizing multimodal representations by jointly modeling behavioral commonality and sample-specific individuality. It combines a learnable global prior with a sample-adaptive residual through gating, and applies polarity-aware adjustment to enhance informative dimensions while suppressing noisy ones.

\noindent\textbf{Mitigating Unimodal Dominance.} 
Multimodal optimization often suffers from imbalanced gradients, leading to unimodal shortcuts~\cite{peng2022ogm}. Representative approaches alleviate this issue through modality dropout~\cite{neverova2016moddrop} or adaptive reweighting~\cite{Huang_2025_CVPR}. DMC addresses this issue during training by attaching auxiliary prediction heads to the visual and audio branches and aligning their predictions with the schema-conditioned multimodal prediction. This guides each modality branch to learn evidence consistent with the final audiovisual judgment, rather than relying on isolated unimodal shortcuts.

\section{Method}
\label{sec:method}
\begin{figure*}[t]
  \centering
  \includegraphics[width=\textwidth]{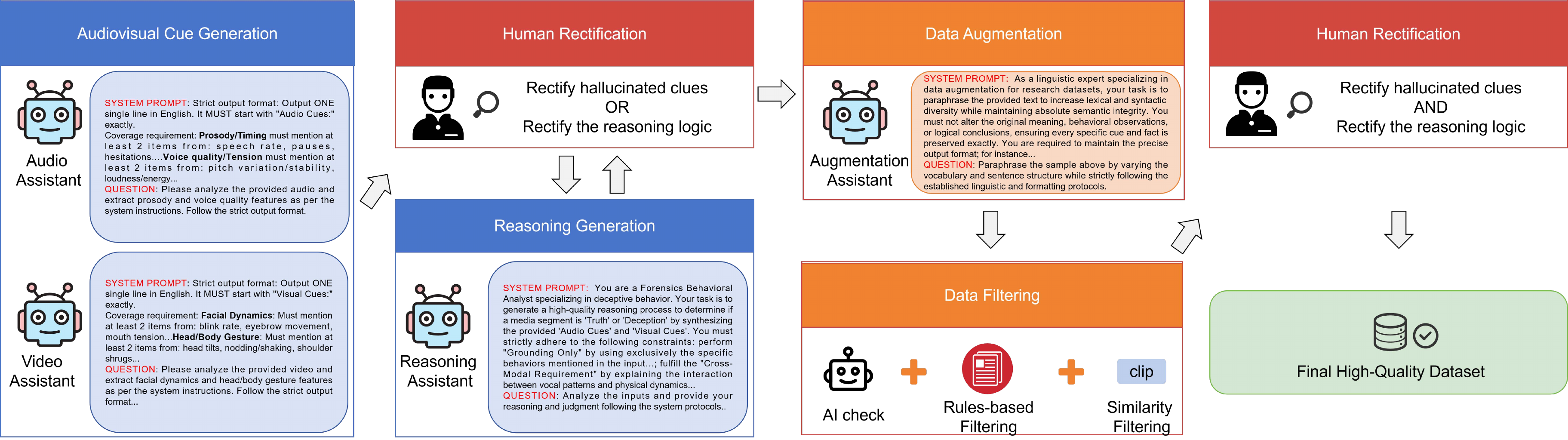}
  \caption{Overview of reasoning dataset generation pipeline. The pipeline adopts a Human-in-the-Loop (HITL) framework to ensure high-quality, auditable structured reports. It begins with AI-driven audiovisual cue extraction, followed by human-guided rectification of hallucinations. A reasoning assistant then synthesizes these cues into forensic judgments. The data is further enriched through semantic augmentation and a multi-tiered filtering stage (comprising AI, rules-based, and CLIP-similarity checks) to produce the final high-fidelity benchmark.
  }
  \label{fig:Data-Generation}
\end{figure*}

\begin{figure}[t]
    \centering
    \begin{subfigure}[b]{0.36\textwidth}
        \centering
        \includegraphics[width=\linewidth]{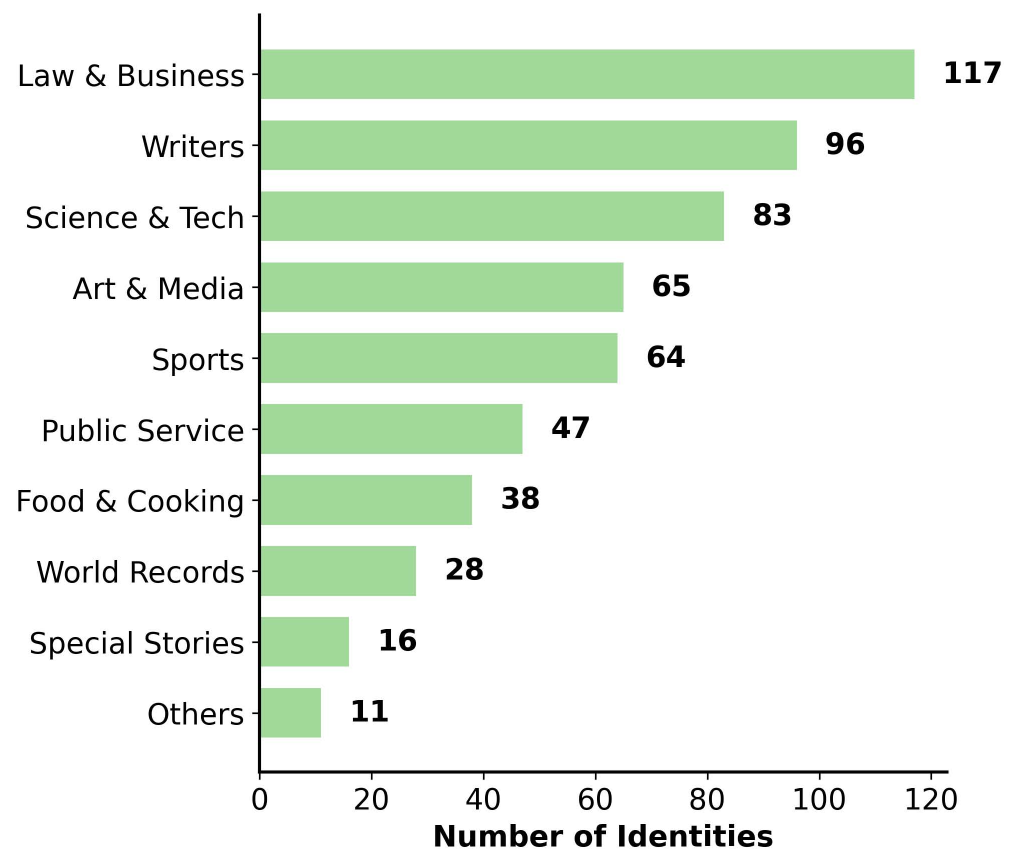}
        \caption{Identity}
    \end{subfigure}
    \hfill
    \begin{subfigure}[b]{0.26\textwidth}
        \centering
        \includegraphics[width=0.9\linewidth]{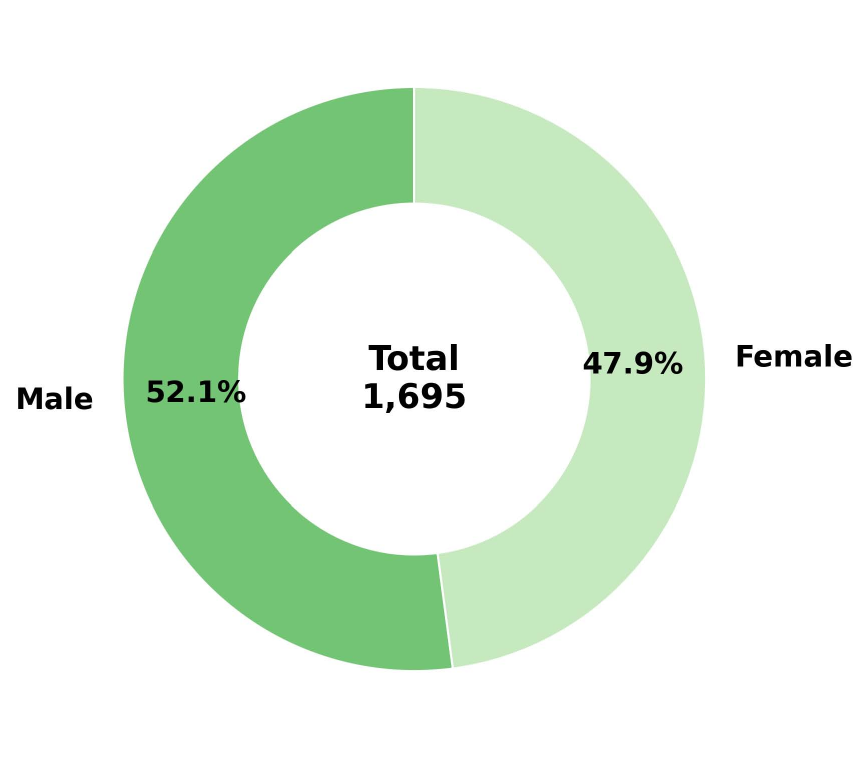}
        \caption{Gender}
    \end{subfigure}
    \hfill
    \begin{subfigure}[b]{0.34\textwidth}
        \centering
        \includegraphics[width=\linewidth]{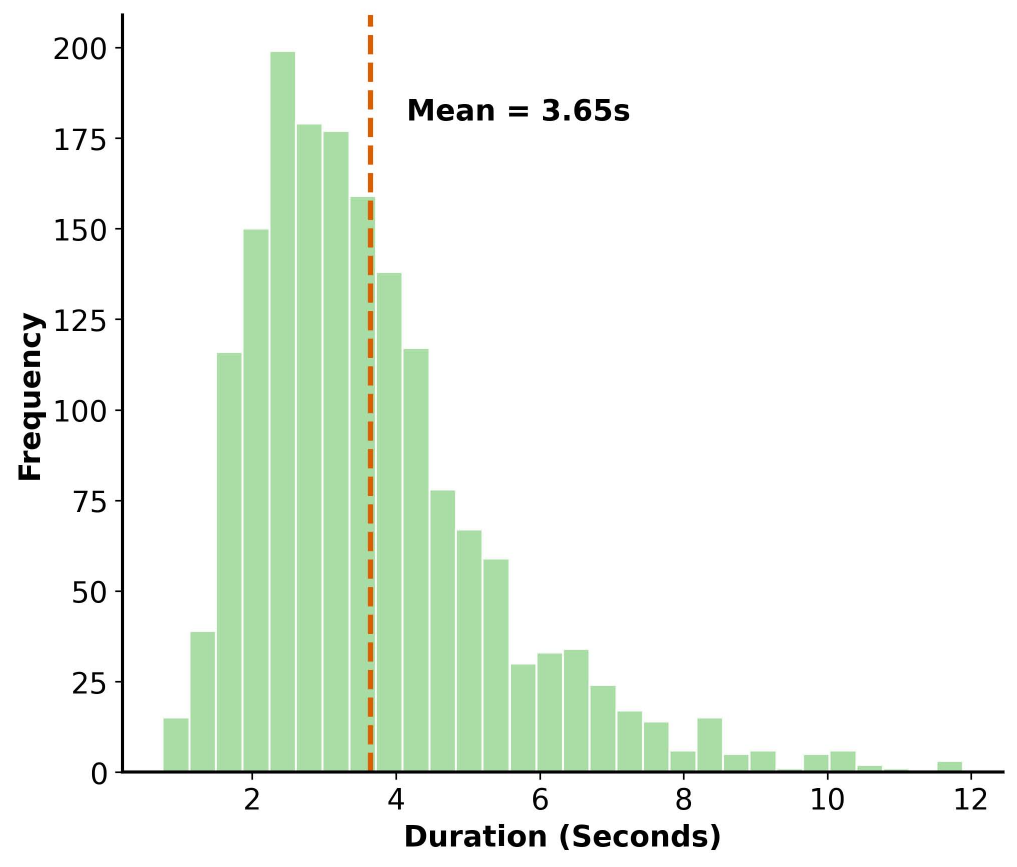}
        \caption{Duration}
    \end{subfigure}

    \caption{Dataset statistics of T4-Deception. We illustrate: (a) distribution of identities, where each one of total 565 identities is shared by one truthful and two deceptive participants; (b) balanced gender distribution; and (c) numerous short-term deceptive segments with an average temporal duration of 3.65s.}
    
    \label{fig:stats}
\end{figure}

\begin{figure*}[t]
  \centering
  \includegraphics[width=\textwidth]{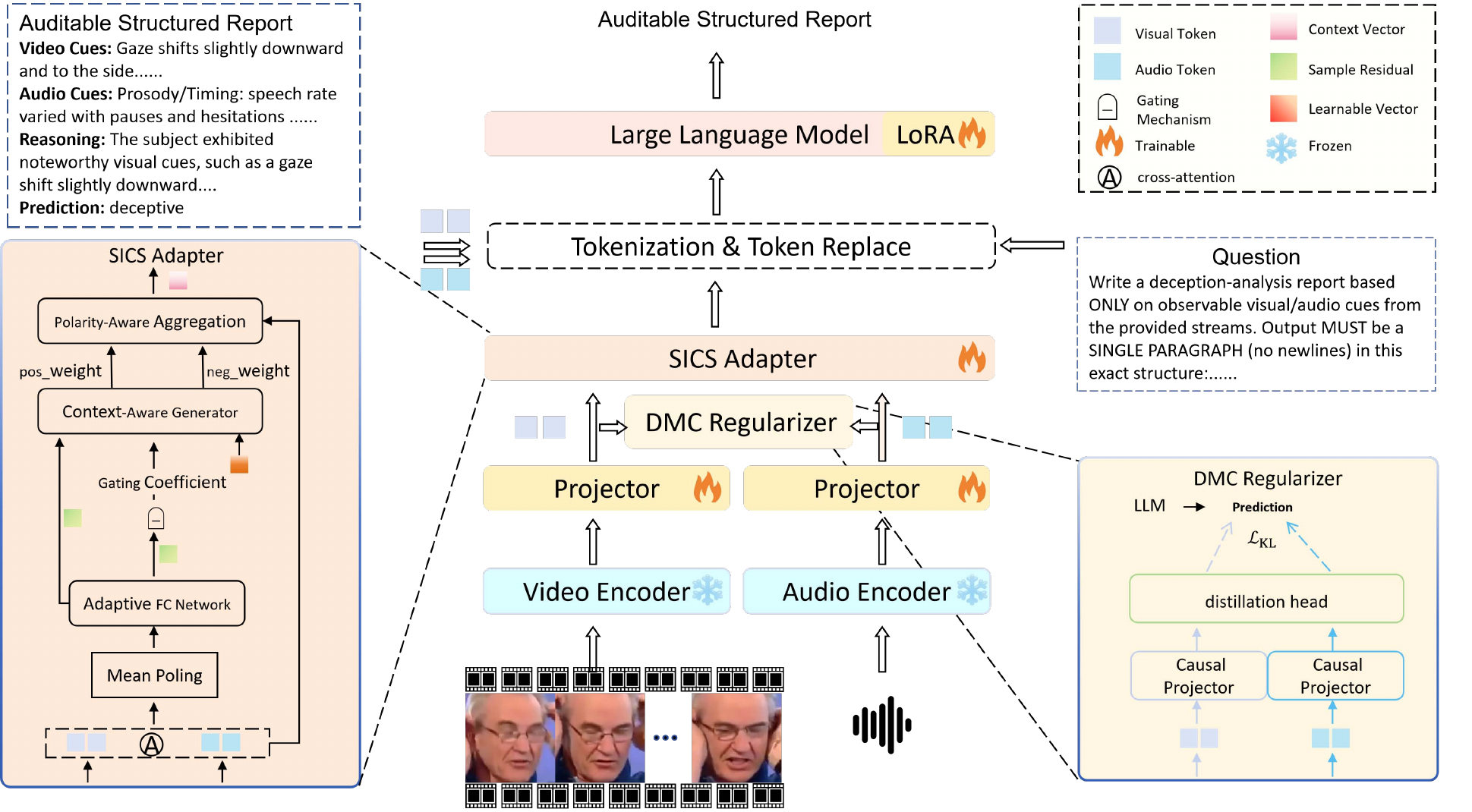}
    \caption{ Overview of our auditable audiovisual deception detection framework. The video and audio encoders extract modality features, which are fused into a robust representation. Within the encoder/fusion stage, SICS stabilizes representations by combining a shared baseline with sample-specific residuals, while DMC mitigates unimodal dominance through agreement regularization on modality-specific prediction distributions. The report generator then produces a single-line schema-constrained audit artifact: \texttt{Video Cues; Audio Cues; Reasoning; Prediction}. }
  \label{fig:teaser}
\end{figure*}

To provide transparency for sensitive decision-making, we propose an auditable MDD framework. As shown in Fig.~\ref{fig:Data-Generation}, we first enrich existing benchmarks with structured behavioral descriptions and reasoning chains. We then introduce T4-Deception (Table~\ref{table:dataset_comparison} and Fig.~\ref{fig:stats}) to cover identity pretense across multicultural scenarios. Finally, we integrate two robust encoding modules (Fig.~\ref{fig:teaser}): SICS, which stabilizes representations by balancing shared behavioral commonalities and sample-specific variations, and DMC, which mitigates unimodal shortcuts through consistency distillation. Together, these components enable schema-constrained audit reports grounded in audiovisual cues.

\begin{table*}[t]
\centering
\caption{Comparison of multimodal deception detection datasets. T4-Deception (To Tell the Truth across 4 cultures) dataset is the largest non-laboratory benchmark, featuring a unified identity pretense task across multiple cultural contexts.}
\label{table:dataset_comparison}
\setlength{\tabcolsep}{4pt} 
\resizebox{\textwidth}{!}{
\begin{tabular}{lccccll}
\toprule
\textbf{Dataset} & \textbf{Subject} & \textbf{Clip} & \textbf{Deceptive} & \textbf{Truthful} & \textbf{Setting} & \textbf{Deceptive Task} \\ \midrule
Real Life Trials \cite{perez2015deception} &56 & 121 & 61 & 60 & Courtroom & False Testimony \\
\midrule
Bag of Lies \cite{gupta2019bagoflies} &35 & 325 & 162 & 163 & Laboratory & False Image-Narration \\
MU3D \cite{lloyd2019miami} &80 & 320 & 160 & 160 & Laboratory & False Social-Evaluation \\ \midrule
Box of Lies \cite{soldner2019box} &26 & 1049 & 862 & 187 & Game Show & False Object-Description \\
DOLOs~\cite{guo2023dolos} &213 & 1675 & 899 & 776 & Game Show & False Story-Telling \\ \midrule
\textbf{T4-Deception (Ours)} &1695 & \textbf{1695} & \textbf{1130} & \textbf{565} & \textbf{Game Show} & \textbf{False Professional-Identity} \\
\hspace{1em}--- \textit{U.S. Edition} &876 & 876 & 584 & 292 & Game Show & (Unified Multi-Culture) \\
\hspace{1em}--- \textit{German Edition} &702 & 702 & 468 & 234 & Game Show & \\
\hspace{1em}--- \textit{Vietnam Edition} &66 &66  & 44 & 22 & Game Show & \\
\hspace{1em}--- \textit{Bulgarian Edition} &51 &51  & 34 & 17 & Game Show & \\ \bottomrule
\end{tabular}%
}
\end{table*}

\subsection{Data Construction: Structured Multi-Cue Reasoning Supervision}
\label{sec:data_construction}

\noindent\textbf{Schema Design.} 
We use a structured output format: \texttt{[Video Cues; Audio Cues; Reasoning; Prediction]}. The model first identifies deception-related audio and visual cues, then performs cross-modal reasoning based on these cues to derive the final prediction. This enforces an evidence-to-conclusion inference flow, where the prediction is derived from explicit behavioral observations through intermediate 
reasoning steps.

\noindent\textbf{HITL Generation Pipeline.} 
As shown in Fig.~\ref{fig:Data-Generation}, our pipeline uses multiple specialized assistants. The Audio Assistant (Qwen-Omni) extracts audio cues such as prosody and speech patterns. The Video Assistant (GPT-4o) extracts visual cues including facial dynamics and body language. A Reasoning Assistant (GPT-4o) then synthesizes these cues into a cross-modal judgment. Human annotators review the outputs from these three assistants to correct hallucinations or logical inconsistencies. An Augmentation Assistant (GPT-4o) then paraphrases the text to increase lexical variation. The augmented samples pass through automated filters before a final human review.

\noindent\textbf{Multi-stage Filtering.} 
After the data augmentation process, we apply three rigorous filtering stages to the augmented samples. First, AI-based checks automatically remove contradictory cue-reasoning pairs to ensure logical consistency. Second, rule-based filtering strictly ensures full format compliance across all entries. Third, similarity filtering effectively prevents data redundancy by eliminating near-duplicate samples. Finally, the filtered samples undergo a comprehensive final human review to confirm overall data quality and reliability.

\noindent\textbf{Annotation Reliability.}
Two annotators performed correction and cross-validation after three pre-annotation calibration rounds. Inter-annotator agreement is computed on discrete cue/reasoning audit labels derived from human-corrected annotations. \textbf{Cohen's $\kappa$ is 0.73}, indicating substantial agreement.

\subsection{T4-Deception (To Tell the Truth across 4 cultures) Dataset}
\label{sec:dataset}

To address cross cultures data scarcity in deception detection, we present T4-Deception dataset, which is also the largest non-laboratory benchmark for multimodal deception detection. T4-Deception studies identity pretense across the United States, Germany, Vietnam, and Bulgaria under a unified game-show format. Detailed source collection, filtering criteria, and label verification protocols are provided in Appendix 1. As detailed in Table~\ref{table:dataset_comparison}, T4-Deception differs from existing game-show datasets that focus on object fabrication (Box of Lies~\cite{soldner2019box}) or story fabrication (DOLOs~\cite{guo2023dolos}). It requires subjects to maintain a fabricated professional identity during interpersonal confrontation, eliciting complex visual and acoustic behavioral markers. With 1,695 samples from four countries and diverse professional identities (Fig.~\ref{fig:stats}), the dataset supports cross-cultural analysis of identity-pretense deception. Its short segments, averaging 3.65s, further encourage models to focus on immediate cues to interpersonal confrontation.

\vspace{-2pt}
\subsection{Stabilized Individuality-Commonality Synergy (SICS)}
\vspace{-1pt}
\label{sec:sics}
The goal of SICS (visualized in the bottom-left inset of Fig. ~\ref{fig:teaser}) is to adapt multimodal features while balancing shared behavioral commonality and sample-specific individuality. In deception detection, global behavioral patterns provide stable cues, whereas each sample may also contain personalized visual-acoustic variations. Therefore, SICS combines a learnable global vector with context-derived residuals to generate sample-adaptive adjustment weights. Positive and negative adjustment branches then enhance useful feature dimensions and suppress noisy or misleading ones, respectively.

Given visual tokens $v_i$ and acoustic tokens $a_i$, we first perform cross-attention between the two modalities to obtain multimodal fusion tokens $x_i \in \mathbb{R}^{L \times d}$, where $L$ denotes the sequence length. The module then calculates the temporal mean of $x_i$ to obtain a context vector $c_i \in \mathbb{R}^d$, which is processed by the adaptive FC network to generate the sample-adaptive residual $\Delta z_i \in \mathbb{R}^d$:

\begin{equation}
\Delta z_i = \mathbf{W}_2 \big(\tanh(\mathbf{W}_1 c_i + \mathbf{b}_1)\big) + \mathbf{b}_2
\end{equation}
where $\mathbf{W}_1, \mathbf{W}_2,  \mathbf{b}_1, \mathbf{b}_2$ are learnable weights and biases.

Subsequently, a gating coefficient $g_i$ is computed to determine the fusion ratio between a learnable global vector $\mathbf{b}_{global} \in \mathbb{R}^d$ and the generated residual $\Delta z_i$: $g_i = \sigma(\mathbf{W}_g \Delta z_i + b_g)$. The Context-Aware Generator then produces positive and negative adjustment weights as follows:
\begin{equation}
\label{eq:refine_weights}
w_i = \tanh\big(g_i \cdot \mathbf{b}_{global} + (1 - g_i) \cdot \Delta z_i\big), \quad w_i^+ = \mathbf{W}^+ w_i + \mathbf{b}^+, \quad w_i^- = \mathbf{W}^- w_i + \mathbf{b}^-
\end{equation}
where $\mathbf{W}^+$, $\mathbf{W}^-$, $\mathbf{b}^+$, and $\mathbf{b}^-$ are independent learnable parameters.

During polarity-aware aggregation, the input features are refined as: $x'_{i} = x_i \odot \operatorname{ReLU}(w_i^+) - x_i \odot \operatorname{ReLU}(w_i^-)$, where $\odot$ denotes element-wise multiplication. Then, the output is calculated as a weighted sum of the refined features and the original input:
\begin{equation}
\label{eq:output_final}
\text{Output}_i = \lambda \cdot x'_{i} + (1 - \lambda) \cdot x_i
\end{equation}
where $\lambda$ is a balancing hyperparameter and we empirically set $\lambda = 0.2$.

\subsection{Distilled Modality Consistency (DMC)}
\label{sec:dmc}
To mitigate unimodal shortcut learning, we introduce the DMC regularizer (visualized in the bottom-right inset of Fig.~\ref{fig:teaser}), which encourages visual and audio modalities to produce consistent predictions during training.

As shown in Fig.~\ref{fig:teaser}, the DMC module consists of modality-specific Causal Projectors followed by a shared distillation head. These components map the frozen visual and audio tokens to the label space 
$\mathcal{Y}=\{\texttt{deceptive}, \texttt{truthful}\}$. Specifically, the unimodal prediction $p_m$ is generated as:
\begin{equation}
\label{eq:mod-head}
p_m(y) = \mathrm{softmax}\big(\Phi_{\mathrm{distill}}(\mathrm{Proj}_m(h_m))\big)(y), 
\quad m \in \{v, a\}, \; y \in \mathcal{Y},
\end{equation}
where $h_m$ represents the modality tokens, $\mathrm{Proj}_m$ is the Causal Projector, and $\Phi_{\mathrm{distill}}$ denotes the distillation head.

The MLLM decoder produces a teacher distribution $q(y)$ at the final \texttt{Prediction} position when conditioning on the full multimodal context and the schema constraint. We minimize the KL divergence between the unimodal predictions from the shared distillation head and this MLLM decoder teacher:

\begin{equation}
\label{eq:mc-distill}
\mathcal{L}_{\mathrm{distill}} = \sum_{m \in \{v, a\}} \mathrm{KL}\big(q \,\|\, p_m\big).
\end{equation}
DMC regularizer encourages both modality projection heads to function effectively, which is discarded during inference.

\subsection{Auditable Report Generation with Schema Constraint}
\label{sec:auditable}

We generate the auditable report directly with an MLLM.
Given video $V_i$ and audio $A_i$, the MLLM encoder produces modality-specific hidden states
and a fused audiovisual representation:
\begin{equation}
\label{eq:mllm-enc}
H_{v,i},\, H_{a,i},\, H_{va,i} = \mathrm{Enc}_{\mathrm{MLLM}}(V_i, A_i),
\end{equation}
where $H_{v,i}$ and $H_{a,i}$ capture modality-specific features for visual and acoustic signals, respectively. $H_{va,i}$ captures the cross-modal interactions between them. 

Conditioned on these representations, the MLLM decoder then generates a structured, single-line schema-constrained report that provides the essential behavioral evidence required for a comprehensive and auditable deception analysis:

\begin{equation}
\label{eq:report}
R_i = \mathrm{Dec}_{\mathrm{MLLM}}\big(H_{v,i},\, H_{a,i},\, H_{va,i};\,\texttt{schema}\big).
\end{equation}
The schema enforces a fixed field order and semicolon delimiters:
\texttt{Video Cues; Audio Cues; Reasoning; Prediction}.
We supervise the MLLM to match the target report text, and require the final \texttt{Prediction} field to match the ground-truth label (\texttt{deceptive}/\texttt{truthful}).

\subsection{Training Objective}
\label{sec:objective}

We jointly optimize the schema-constrained report generation task alongside regularizers. The overall objective function is defined as:
\begin{equation}
\label{eq:overall}
\mathcal{L} = \mathcal{L}_{\mathrm{rep}} + \alpha \mathcal{L}_{\mathrm{distill}},
\end{equation}
where $\mathcal{L}_{\mathrm{rep}}$ denotes the token-level cross-entropy loss for generating the structured audit report. The term $\mathcal{L}_{\mathrm{distill}}$ represents the consistency distillation loss derived from our DMC module. We empirically set $\alpha = 0.1$ to balance the accuracy of the generated reports and the mitigation of unimodality reliance.

\section{Experiments}
\label{sec:experiments}

We evaluate our method on six aspects: (i) in-domain effectiveness across diverse deceptive tasks, (ii) cross-domain generalization under dataset shift, (iii) cross-cultural robustness in identity pretense, (iv) component contributions via ablation studies, (v) visualization analysis of SICS adapter and DMC regularizer, (vi) reasoning capability analysis.

\subsection{Implementation Details}
\label{sec:exp_impl}
We use AffectGPT~\cite{lian2024affectgpt} as the base MLLM, which consists of visual/audio encoders, audiovisual projection modules, and an LLM decoder, initialized with the corresponding pre-trained weights. During fine-tuning, the encoders remain frozen; we apply LoRA to the LLM while fully training the projectors and our proposed components. Optimization is performed for 200 epochs using AdamW\cite{Loshchilov2017DecoupledWD} ($LR=5\times10^{-5}$). Training is conducted on a single NVIDIA H100 (80GB) with a batch size of 4, sampling 8 frames per video at $224 \times 224$ resolution.

\subsection{Datasets and Protocols}
\label{sec:exp_data_protocol}
To verify the effectiveness of our method, we evaluate our model on three established benchmarks and our newly introduced dataset, T4-Deception:
\begin{itemize}
\item \textbf{Bag-of-Lies (BoL)}~\cite{gupta2019bagoflies}: 325 lab-collected samples on false image-narration.
\item \textbf{MU3D}~\cite{lloyd2019miami}: 320 lab-collected samples on false social-evaluation.
\item \textbf{DOLOs}~\cite{guo2023dolos}: 1,675 game show samples on false story-telling.
\item \textbf{T4-Deception (Ours)}: 1,695 samples across four cultural contexts:  U.S. (876), Germany (702), Vietnam (66), and Bulgaria (51), collected under a unified false professional-identity task.
\end{itemize}

For the established benchmarks, we conduct in-domain evaluations following official protocols (3-fold for BoL and DOLOs; 4-fold for MU3D), alongside cross-domain assessments. We also perform in-cultural evaluations via 3-fold cross-validation and pairwise cross-cultural tests.

\subsection{MLLM Configuration}
\label{sec:exp_setup}

We compare our method against two types of MLLMs:

\noindent\textbf{Commercial MLLMs (Zero-shot).}
Commercial models, including GPT-4o~\cite{openai2024gpt4o} and so on, are evaluated via API calls in a zero-shot setting due to limited parameter access. These represent the reasoning capabilities of general-purpose models without task-specific training.

\noindent\textbf{Open-source MLLMs (Fine-tuned).}
We fine-tune representative open-source MLLMs, including Qwen3-Omni~\cite{qwen25omni2025}, VideoLLaMA2~\cite{videollama2} and so on. These models are trained using LoRA on the same training sets as our method. 

\noindent\textbf{Unified Prompting and Output Schema.}
All models use the same schema-constrained prompt format. Models output a single-line structured record: \texttt{Video Cues; Audio Cues; Reasoning; Prediction}. 

\subsection{In-Domain Evaluation}
\label{sec:exp_indomain}

\begin{table*}[htpb]
\centering
\caption{Comprehensive In-domain Benchmark. We evaluate performance across three datasets: DOLOs, Bag-of-Lies (BoL), and MU3D. \textbf{Set.}: ZS=Zero-shot, LoRA=Fine-tuning, Full=Full Training. \textbf{Mod.}: A=Audio, V=Video, AV=Audio-Visual.}
\label{tab:final_benchmark}
\scriptsize 
\setlength{\tabcolsep}{0pt} 
\renewcommand{\arraystretch}{1.2}
\begin{tabular*}{\textwidth}{@{\extracolsep{\fill}} l l l c cc cc cc @{}}
\toprule
& & & & \multicolumn{2}{c}{\textbf{DOLOs}} & \multicolumn{2}{c}{\textbf{BoL}} & \multicolumn{2}{c}{\textbf{MU3D}} \\
\cmidrule(lr){5-6} \cmidrule(lr){7-8} \cmidrule(lr){9-10}
\textbf{Method/Model} & \textbf{Source} & \textbf{Set.} & \textbf{Mod.} & \textbf{Acc.} & \textbf{F1} & \textbf{Acc.} & \textbf{F1} & \textbf{Acc.} & \textbf{F1} \\
\midrule
\rowcolor[gray]{.95} \multicolumn{10}{l}{\textit{Task-specific Deep Learning Methods}} \\
LieNet~\cite{karnati2022lienet} & TCDS'22 & Full & AV & 56.50 & 69.72 & 59.78 & 58.14 & 53.48 & 33.62 \\
FacialCueNet~\cite{nam2023facialcuenet} & AI'23 & Full & V & 60.98 & 68.65 & 56.23 & 63.26 & 57.64 & 59.13 \\
PECL~\cite{guo2023dolos} & ICCV'23 & Full & AV & 64.75 & 71.20 & 59.51 & 51.06 & 55.31 & 60.07 \\
AFFAKT~\cite{Ji2025AFFAKTAH} & AAAI'25 & Full & AV & 68.10 & 70.73 & -- & -- & -- & -- \\
\midrule
\rowcolor[gray]{.95} \multicolumn{10}{l}{\textit{Large Multimodal Models (LMMs)}} \\
GPT-4o~\cite{openai2024gpt4o}    & Comm. & ZS & V  & 66.38 & 64.21 & 57.14 & 56.35 & 54.12 & 52.47 \\
Qwen-Omni~\cite{qwen25omni2025}  & Comm. & ZS & AV & 51.72 & 49.56 & 45.27 & 35.82 & 50.23 & 41.38 \\
VideoLLaMA2~\cite{videollama2}   & Open & LoRA & AV & 53.48 & 56.12 & 47.65 & 49.34 & 51.84 & 54.62 \\
SALMONN2~\cite{Tang2025videoSALMONN2C}    & Open & LoRA & AV & 52.63 & 55.44 & 46.82 & 48.51 & 51.06 & 53.79 \\
Qwen2.5-7B-VL~\cite{qwen25omni2025} & Open & LoRA & AV & 46.10 & 52.37 & 44.24 & 41.15 & 53.86 & 56.24 \\
\midrule
\textbf{DecepGPT} & \textbf{ours} & \textbf{LoRA} & \textbf{AV} & \textbf{73.23} & \textbf{76.13} & \textbf{63.46} & \textbf{63.72} & \textbf{61.25} & \textbf{67.22} \\
\bottomrule
\end{tabular*}
\end{table*}

As shown in Table~\ref{tab:final_benchmark}, our method achieves state-of-the-art performance across all three benchmarks. First, compared to zero-shot commercial giants (e.g., GPT-4o~\cite{openai2024gpt4o}), our method yields consistent accuracy gains of +6.85\% on DOLOs, +6.32\% on BoL, and +7.13\% on MU3D. Second, compared to fine-tuned open-source MLLMs (e.g., VideoLLaMA2~\cite{videollama2}), our method yields consistent accuracy gains of +19.75\% on DOLOs, +15.81\% on BoL, and +7.39\% on MU3D. Finally, our approach even surpasses specialized task-specific models (e.g., PECL~\cite{guo2023dolos}, AFFAKT~\cite{Ji2025AFFAKTAH}) by margins of +5.13\% (DOLOs), +3.68\% (BoL), and +3.61\% (MU3D). These results are achieved via efficient LoRA fine-tuning while generating structured audit reports, offering both higher accuracy and better interpretability than baselines that output simple labels. Additional results on in-the-wild dataset are provided in Appendix 2. Beyond classification accuracy, we rigorously evaluate the quality of the generated auditable reports in Table~\ref{tab:language_metrics}. DecepGPT achieves superior performance across all metrics compared to general-purpose MLLMs. The high BERTScore (0.812) and GPT-Score (6.96) indicate that our schema-driven approach effectively grounds the reasoning in forensic evidence rather than generating generic hallucinations.

\begin{table}[t]
\centering
\caption{Linguistic evaluation of generated reasoning reports. Higher scores indicate better quality.}
\label{tab:language_metrics}
\resizebox{0.8\textwidth}{!}{%
\begin{tabular}{l cccc}
\toprule
\textbf{Method} & \textbf{METEOR} & \textbf{ROUGE-L} & \textbf{BERTScore} & \textbf{GPT-Score} \\
\midrule
GPT-4o (Zero-shot)  & 0.284 & 0.312 & 0.765 & 5.42 \\
VideoLLaMA2 (LoRA)  & 0.215 & 0.246 & 0.792 & 5.15 \\
Qwen2.5-7B-VL (LoRA)& 0.242 & 0.285 & 0.804 & 5.28 \\
\midrule
\rowcolor{gray!10} \textbf{DecepGPT} & \textbf{0.342} & \textbf{0.384} & \textbf{0.812} & \textbf{6.96} \\
\bottomrule
\end{tabular}%
}
\end{table}

\subsection{Cross-Domain Evaluation}
\label{sec:exp_crossdomain}
We evaluate our method under dataset shift by training on source datasets and testing on target domains without target-domain fine-tuning. Table~\ref{tab:fusion_results_clean} demonstrates that our method consistently outperforms task-specific methods (e.g., PECL) on all transfer paths in accuracy: +8.95\% (M\&B $\to$ D), +4.93\% (D\&M $\to$ B), and +1.86\% (D\&B $\to$ M), with a +6.40\% average gain. Regarding F1 score, our method also achieves improvements: +5.05\% (D\&M $\to$ B) and +0.63\% (D\&B $\to$ M), with a +1.73\% average gain. 

\begin{table}[t]
\centering
\caption{Cross-Domain Evaluation on DOLOs (D), Bag-of-Lies (B), and MU3D (M). ``X\&Y$\rightarrow$Z'' denotes training on X and Y, testing on Z.}
\label{tab:fusion_results_clean}
\renewcommand{\arraystretch}{1.2} 
\setlength{\tabcolsep}{4pt} 
\begin{tabular}{l cccccccc}
\toprule 
\multirow{2}{*}{Method} & \multicolumn{2}{c}{M\&B $\rightarrow$ D} & \multicolumn{2}{c}{D\&M $\rightarrow$ B} & \multicolumn{2}{c}{D\&B $\rightarrow$ M} & \multicolumn{2}{c}{Average} \\ 
\cmidrule(lr){2-3} \cmidrule(lr){4-5} \cmidrule(lr){6-7} \cmidrule(lr){8-9} 
 & Acc. & F1 & Acc. & F1 & Acc. & F1 & Acc. & F1 \\ 
\midrule 
LieNet~\cite{karnati2022lienet} & 54.40 & 68.23 & 54.69 & 50.51 & 51.08 & 59.75 & 53.39 & 59.50 \\
PECL~\cite{guo2023dolos}        & 54.51 & \textbf{69.55} & 51.25 & 66.95 & 55.38 & 52.46 & 53.71 & 62.99 \\
\textbf{Ours} & \textbf{63.46} & 61.79 & \textbf{59.62} & \textbf{72.00} & \textbf{57.24} & \textbf{60.38} & \textbf{60.11} & \textbf{64.72} \\ 
\bottomrule 
\end{tabular}
\end{table}

\subsection{Evaluation on T4-Deception}
\label{sec:exp_crosscultural}

We conduct a comprehensive evaluation on the T4-Deception dataset, including both \textbf{in-cultural} and \textbf{cross-cultural} assessments across four distinct regions: U.S., Germany, Vietnam, and Bulgaria.
As shown in Table~\ref{tab:cross_cultural_matrix}, we train models on each source culture and evaluate them on all four targets (including the source itself) in a zero-shot manner.
In-cultural Performance: When trained and tested on the same culture (diagonal entries), our method achieves stable baseline performance, with an average accuracy of 62.58\% (e.g., 64.65\% for U.S. and 61.11\% for Germany). This confirms the model's ability to capture culture-specific deception markers effectively.
More importantly, the model demonstrates strong generalization capabilities when transferred to unseen cultures (off-diagonal entries). For instance, the model trained on the U.S. region retains 59.50\% accuracy when tested on Germany, showing only a marginal drop compared to the in-domain setting. Across all 12 cross-cultural transfer pairs, the average performance remains stable at 57.69\%, with an average relative degradation of only 4.89\% compared to in-cultural results.  Further comparative analysis of cross-cultural deception is provided in Appendix 1.

\begin{table*}[t]
\centering
\caption{Comprehensive Evaluation on T4-Deception. We report Accuracy (Acc) and F1 Score for both in-cultural (diagonal, \textbf{bold}) and cross-cultural (off-diagonal) settings. Models are trained on the row culture and tested on the column culture.}
\label{tab:cross_cultural_matrix}
\small
\setlength{\tabcolsep}{3.5pt} 
\renewcommand{\arraystretch}{1.2}
\resizebox{0.9\textwidth}{!}{%
\begin{tabular}{l cc cc cc cc}
\toprule
\multirow{2}{*}{\textbf{Train $\downarrow$ / Test $\to$}} & \multicolumn{2}{c}{\textbf{U.S.}} & \multicolumn{2}{c}{\textbf{Germany}} & \multicolumn{2}{c}{\textbf{Vietnam}} & \multicolumn{2}{c}{\textbf{Bulgaria}} \\
\cmidrule(lr){2-3} \cmidrule(lr){4-5} \cmidrule(lr){6-7} \cmidrule(lr){8-9}
& \textbf{Acc} & \textbf{F1} & \textbf{Acc} & \textbf{F1} & \textbf{Acc} & \textbf{F1} & \textbf{Acc} & \textbf{F1} \\ \midrule
\textbf{U.S.} & \textbf{64.65} & \textbf{73.28} & 59.50 & 52.14 & 57.58 & 61.20 & 63.00 & 71.62 \\
\textbf{Germany} & 57.00 & 57.26 & \textbf{61.11} & \textbf{53.33} & 56.06 & 59.14 & 60.78 & 62.45 \\
\textbf{Vietnam} & 55.48 & 61.05 & 54.87 & 48.92 & \textbf{63.10} & \textbf{68.42} & 58.82 & 64.10 \\
\textbf{Bulgaria} & 58.19 & 63.38 & 56.45 & 50.18 & 54.55 & 58.76 & \textbf{61.46} & \textbf{66.25} \\ \bottomrule
\end{tabular}%
}
\end{table*}

\subsection{Ablation Study}
\label{sec:ablation}

\noindent\textbf{Main Ablation Study.}
We evaluate the contribution of each core component across \textbf{all three datasets} in Table~\ref{tab:ablation_all}. 
Both SICS and DMC consistently improve over the baseline. 
On DOLOs, SICS provides a significant accuracy gain of +5.44\%, while DMC adds a further +1.53\%. 
Similarly, on Bag-of-Lies, SICS boosts accuracy by +4.81\%, with DMC contributing an additional +1.93\%. 
For MU3D, we observe a substantial improvement of +6.40\% from SICS and +1.25\% from DMC. 
The full model, combining both modules, achieves the best performance across all benchmarks, yielding an average accuracy gain of +6.42\% over the baseline. 
These results confirm that SICS and DMC address complementary aspects of robust multimodal learning. The cross-domain transfer performance of these ablated variants is provided in Appendix 3.

\begin{table}[t]
\centering
\caption{Main ablation study across three datasets. Base refers to the model without the SICS adapter and DMC regularization.}
\label{tab:ablation_all}
\setlength{\tabcolsep}{3.5pt} 
\renewcommand{\arraystretch}{1.2}
\small
\begin{tabular}{l cc cc cc}
\toprule
& \multicolumn{2}{c}{\textbf{DOLOs}} & \multicolumn{2}{c}{\textbf{Bag-of-Lies}} & \multicolumn{2}{c}{\textbf{MU3D}} \\
\cmidrule(lr){2-3} \cmidrule(lr){4-5} \cmidrule(lr){6-7} 
\textbf{Variant} & \textbf{Acc.} & \textbf{F1} & \textbf{Acc.} & \textbf{F1} & \textbf{Acc.} & \textbf{F1}\\
\midrule
Base (no SICS/DMC) & 67.24 & 71.18 & 57.69 & 54.26 & 53.75 & 65.34 \\
Base + DMC         & 68.77 & 71.94 & 59.62 & 58.15 & 55.00 & 66.27 \\
Base + SICS        & 72.68 & 76.02 & 62.50 & 61.43 & 60.15 & 67.08 \\
\midrule
\textbf{Full (SICS + DMC)} & \textbf{73.23} & \textbf{76.13} & \textbf{63.46} & \textbf{63.72} & \textbf{61.25} & \textbf{67.22} \\
\bottomrule
\end{tabular}
\end{table}

\noindent\textbf{Component Analysis and Backbone Comparison.}
To further validate the effectiveness of the internal mechanisms, we conduct detailed ablation studies on the SICS adapter components and compare different backbone networks. 
Due to space limitations, we report the results on the \textbf{DOLOs} dataset here as a representative example; the complete results for Bag-of-Lies and MU3D are detailed in Appendix 4. As shown in Table~\ref{tab:sics_ablation}, removing any component of the SICS adapter (global prior, polarity-aware adjustment, or gating mechanism) leads to a clear drop in performance, confirming their necessity. 
Furthermore, we investigate the impact of the backbone architecture (Table~\ref{tab:backbone_comp}). Replacing the LLM backbone with simpler structures (MLP or Transformer) results in significantly lower performance. This suggests that the LLM backbone provides superior semantic reasoning capabilities essential for this task.

\begin{table}[t]
\centering
\captionof{table}{Ablation on SICS components.}
\label{tab:sics_ablation}
\renewcommand{\arraystretch}{0.9}
\resizebox{\linewidth}{!}{%
\begin{tabular}{@{}l|cccc|ccc|c@{}}
\toprule
\textbf{Variant} & Base & +Glo. & +Pol. & +Gat. & +Glo.\&Pol. & +Glo.\&Gat. & +Pol.\&Gat. & \textbf{Full} \\ \midrule
\textbf{Acc. (\%)} & 67.24 & 68.45 & 68.81 & 68.62 & 71.85 & 72.06 & 71.42 & \textbf{73.23} \\
\textbf{F1 Score}  & 71.18 & 72.03 & 72.45 & 72.16 & 74.94 & 75.28 & 74.31 & \textbf{76.13} \\ \bottomrule
\end{tabular}}
\end{table}

\begin{figure*}[t]
  \centering

  \begin{minipage}[c]{0.28\textwidth}
    \centering
    \scriptsize
    \renewcommand{\arraystretch}{1.0}
    \setlength{\tabcolsep}{3pt}
    \captionof{table}{Backbone comparison.}
    \label{tab:backbone_comp}
    \resizebox{\linewidth}{!}{%
    \begin{tabular}{l cc}
      \toprule
      \textbf{Backbone} & \textbf{Acc.} & \textbf{F1} \\ 
      \midrule
      MLP~\cite{tolstikhin2021mlp} & 58.42 & 60.17 \\
      Trans.~\cite{vaswani2017attention} & 64.75 & 68.34 \\ 
      \midrule
      \textbf{Ours} & \textbf{73.23} & \textbf{76.13} \\ 
      \bottomrule
    \end{tabular}%
    }
  \end{minipage}
  \hfill
  \begin{minipage}[c]{0.7\textwidth}
    \centering
    \begin{subfigure}[b]{0.48\linewidth}
      \centering
      \includegraphics[width=\linewidth]{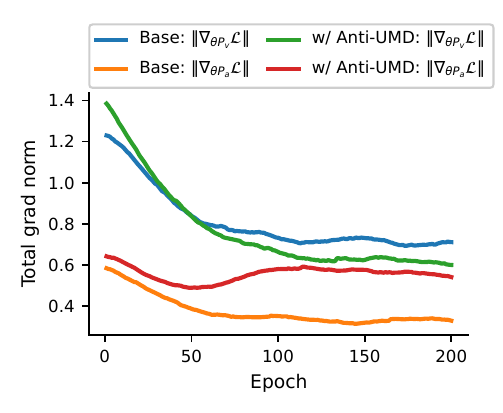}
      \caption{Gradient norm.}
      \label{fig:grad_balance}
    \end{subfigure}
    \hfill
    \begin{subfigure}[b]{0.48\linewidth}
      \centering
      \includegraphics[width=\linewidth]{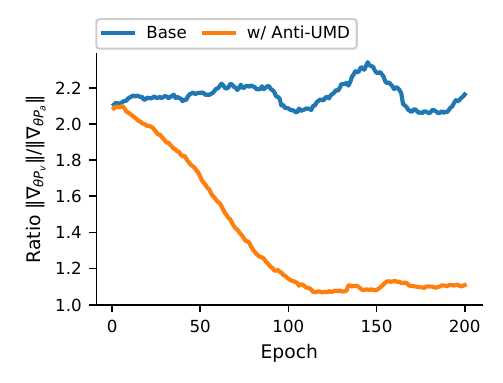}
      \caption{Gradient ratio.}
      \label{fig:grad_ratio}
    \end{subfigure}

    \caption{Projector gradient dynamics analysis during training.}
    \label{fig:combined_grad_analysis}
  \end{minipage}

\end{figure*}

\section{Visualization and Interpretability Analysis}
\label{sec:visualization}

We analyze the internal mechanisms of the SICS Adapter and DMC Regularizer from two aspects: (i) unimodal shortcut mitigation in the DMC Regularizer and (ii) representation stabilization in the SICS Adapter. We also assess the fidelity of the generated auditable evidence, verifying that it is grounded in the extracted behavioral cues to provide a transparent audit trail for the final prediction.

\begin{figure*}[t]
\centering
\begin{subfigure}[b]{0.32\textwidth}
\centering
\includegraphics[width=\linewidth]{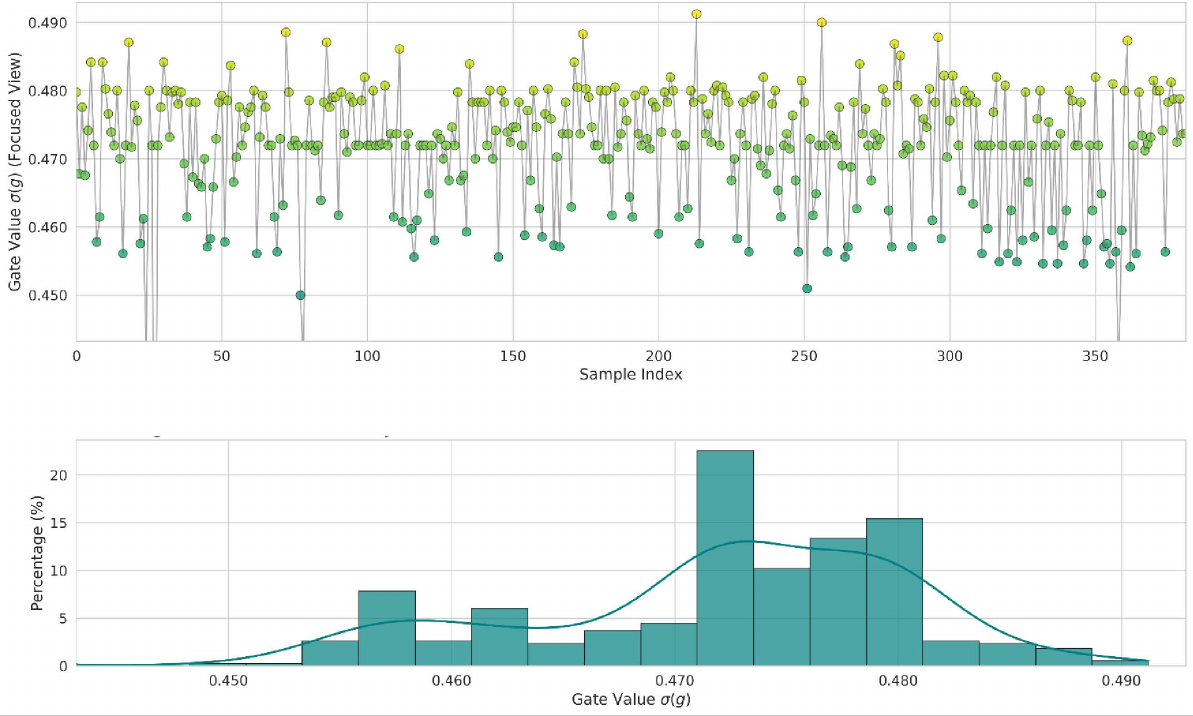}
\caption{Gating distribution $\sigma(g)$.}
\label{fig:gating}
\end{subfigure}
\hfill
\begin{subfigure}[b]{0.32\textwidth}
\centering
\includegraphics[width=\linewidth]{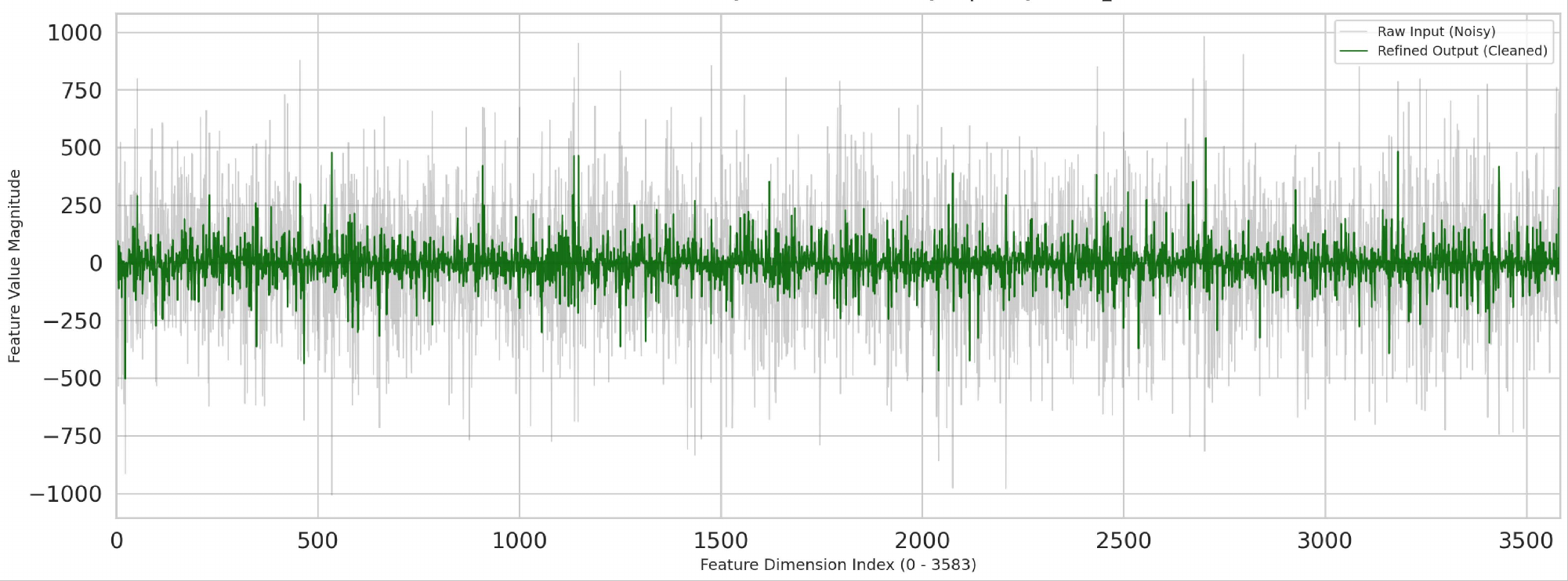}
\caption{SICS feature stabilization.}
\label{fig:denoising}
\end{subfigure}
\hfill
\begin{subfigure}[b]{0.32\textwidth}
\centering
\includegraphics[width=\linewidth]{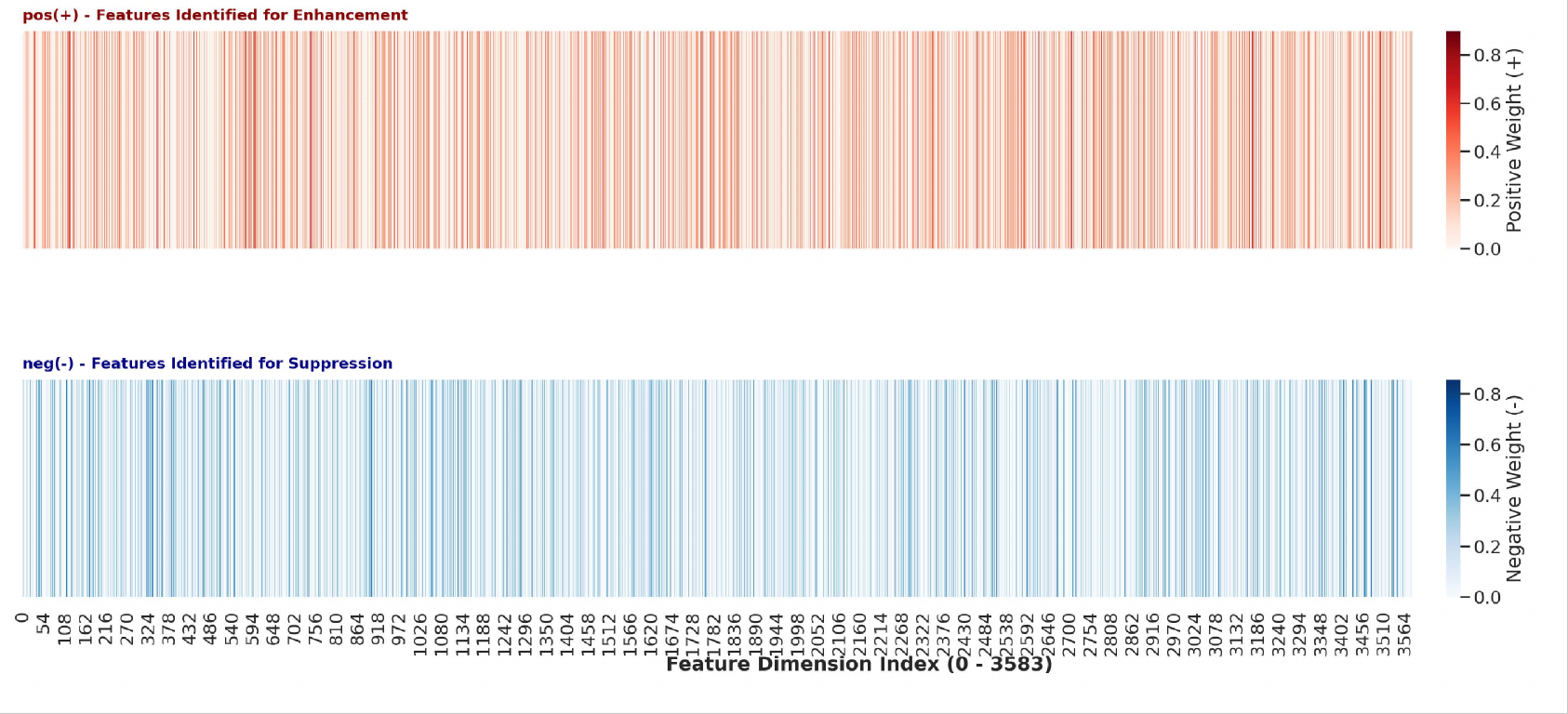}
\caption{Polarity-aware weights.}
\label{fig:causal_split}
\end{subfigure}
\caption{Visualization analysis of the SICS Adapter. \textbf{(a)} Dynamic variation of gating values on the fusion of global priors and local residuals. \textbf{(b)} Comparison between volatile raw features and stabilized features processed by SICS. \textbf{(c)} Adjustment where positive weights enhance informative feature and negative weights suppress noise features.}
\label{fig:mechanisms_analysis}
\end{figure*}

\begin{figure}[t]
  \centering
  \begin{subfigure}[b]{0.56\textwidth}
    \centering
    \includegraphics[width=\linewidth]{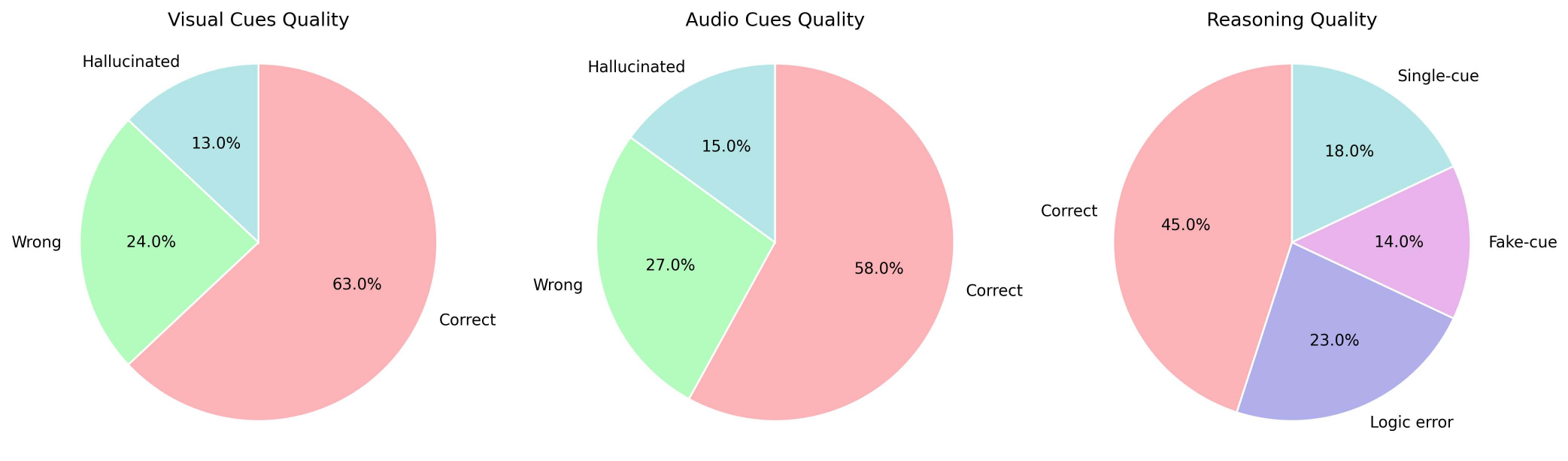}
    \caption{Audit taxonomy distribution.}
    \label{fig:audit_pies}
  \end{subfigure}
  \hfill
  \begin{subfigure}[b]{0.42\textwidth}
    \centering
    \includegraphics[width=\linewidth]{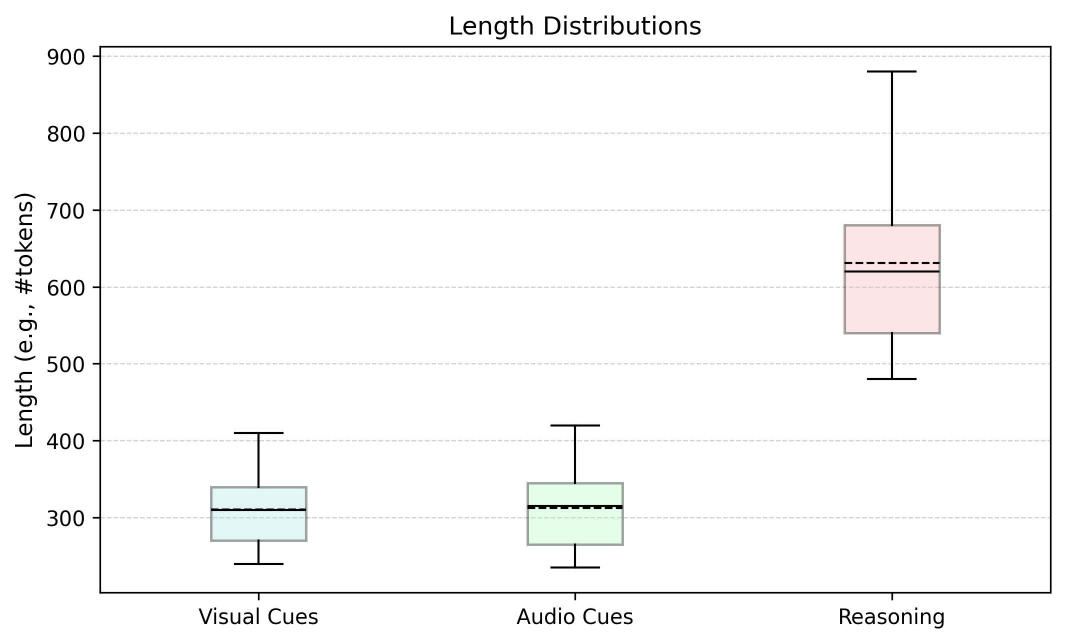}
    \caption{Response length distributions.}
    \label{fig:audit_lengths}
  \end{subfigure}
    \caption{Quantitative audit analysis. (a) distribution of cue correctness and 
    reasoning quality categories; (b) text length distribution of reasoning.}
  \label{fig:audit_combined}
\end{figure}

\subsection{Visualization Analysis of the DMC Regularizer}
\label{sec:viz_DMC}
Fig.~\ref{fig:combined_grad_analysis} illustrates how the DMC Regularizer affects cross-modal feature utilization. Fig.~\ref{fig:combined_grad_analysis}(a) tracks the gradient norms backpropagated to the modality-specific projectors. In the base setting, the visual projector receives larger gradients, indicating a bias toward visual cues. With DMC, the acoustic-side gradients are gradually strengthened, suggesting that the model incorporates more audio evidence during training. Fig.~\ref{fig:combined_grad_analysis}(b) shows the dynamics via the ratio $r = \|\nabla_{\theta P_v}\| / \|\nabla_{\theta P_a}\|$. The base model shows visual dominance ($r > 1$), while the DMC Regularizer reduces this ratio closer to unity. These results indicate that the DMC Regularizer helps balance gradient flow across modalities, supporting more balanced multimodal learning.

\subsection{Visualization Analysis of the SICS Adapter}
\label{sec:viz_SICS}
We visualize the adaptive gating distribution, the denoising effect, and the polarity-aware weights to illustrate the SICS Adapter as shown in Fig. \ref{fig:mechanisms_analysis}.

\noindent\textbf{Dynamic gating behavior.}
Fig.~\ref{fig:gating} shows the gating coefficient $\sigma(g)$ varies across samples (mostly within $[0.443, 0.493]$). This indicates that the fusion is not a static offset but adapts to each sample. This supports the individuality--commonality synergy: the adapter maintains stable features for typical cases while allowing flexibility to model persona-specific residuals for outliers.

\noindent\textbf{Feature stabilization and noise suppression.}
In small-data regimes, audiovisual learning can be affected by high-magnitude persona-driven noise (e.g., individual behaviors or recording conditions). Fig.~\ref{fig:denoising} compares feature magnitudes before and after SICS Adapter: raw features show spiky, volatile dimensions, while refined features are smoother and more centered. This suggests that the SICS Adapter reduces such variations, improving numerical stability.

\noindent\textbf{Polarity-aware enhancement vs. suppression.}
Fig.~\ref{fig:causal_split} visualizes the learned positive (enhancement) and negative (suppression) weights. The adjustment from polarity-aware may help produce more discriminative features for the MLLM.

\subsection{Validity Analysis of Auditable Reasoning}
\label{sec:audit}

We use structured, schema-constrained reports (\texttt{Vi\hspace{0pt}deo Cues};\allowbreak{} \texttt{Au\hspace{0pt}dio Cues};\allowbreak{} \texttt{Rea\hspace{0pt}soning};\allowbreak{} \texttt{Pre\hspace{0pt}diction}) as audit artifacts to identify hallucinations and shortcut rationales. We audit these reports along two axes with an error taxonomy: (i) Visual/Acoustic Cues, categorized as Correct, Counterfactual, or Non-existent (hallucinated); (ii) Reasoning Quality, classified as Correct, False-cue (logical but counterfactual), Incoherent, or Single-cue (modal collapse). Since textual GT descriptions cannot exhaustively cover all valid audio-visual observations, case-by-case expert analysis against the original videos is necessary to avoid penalizing correct but unannotated cues. We randomly sample 100 model outputs for human auditing and further adopt a cascaded taxonomy, where each output is assigned to exactly one category within each axis: cue existence is verified before attribute-level errors for audio-visual cues, and reasoning chains are checked sequentially for multimodal cue usage, logical coherence, and factual consistency. 

Quantitative distributions and answer length statistics (Fig.~\ref{fig:audit_combined}) demonstrate that our framework enhances behavioral understanding. To highlight how the generated schema-constrained reasoning chains reveal the evidentiary basis for each prediction, we also perform qualitative analysis in Appendix 5.

\section{Conclusion} 
\label{sec:conclusion}
We present DecepGPT, an auditable MDD framework targeting evidence-based and generalizable deception detection in high-stakes applications. Through structured cue-reasoning supervision and T4-Deception, DecepGPT supports human-checkable reports and cross-cultural evaluation; through proposed SICS and DMC, it improves robust multimodal learning under small-data conditions. Quantitative experiments and qualitative audits demonstrate that our framework improves detection performance while making final judgments easier to verify.

\section*{Acknowledgements}
This work was supported in part by the National Natural Science Foundation of China (Grant No. 62576076 and No. 62441619), Ningbo Science and Technology Innovation 2025 Major Project (2025Z027), Guangdong Basic and Applied Basic Research Foundation (Grant No. 2023A1515140037), CCF-Tencent Rhino-Bird Open Research Fund, Guangdong Research Team for Communication and Sensing Integrated with Intelligent Computing (Project No. 2024KCXTD047), and SongShan Lake HPC Center (SSL-HPC) in Great Bay University.

%
%
\bibliographystyle{splncs04}
\bibliography{main}

\clearpage
\appendix

\setcounter{section}{0}
\setcounter{subsection}{0}
\setcounter{table}{0}
\setcounter{figure}{0}

\renewcommand{\thesection}{\arabic{section}}
\renewcommand{\thesubsection}{\thesection.\arabic{subsection}}
\renewcommand{\thetable}{\arabic{table}}
\renewcommand{\thefigure}{\arabic{figure}}

\vspace{20pt} 
\begin{center}
    \LARGE \bfseries Appendix
\end{center}
\vspace{10pt}

\vspace{-4pt}
\section{Comprehensive Analysis of T4-Deception}
\label{app:behavioral_analysis}
\vspace{-4pt}

\begin{figure}[htbp]
    \centering
    \begin{minipage}[c]{0.6\linewidth}
        \centering
        \includegraphics[width=\linewidth]{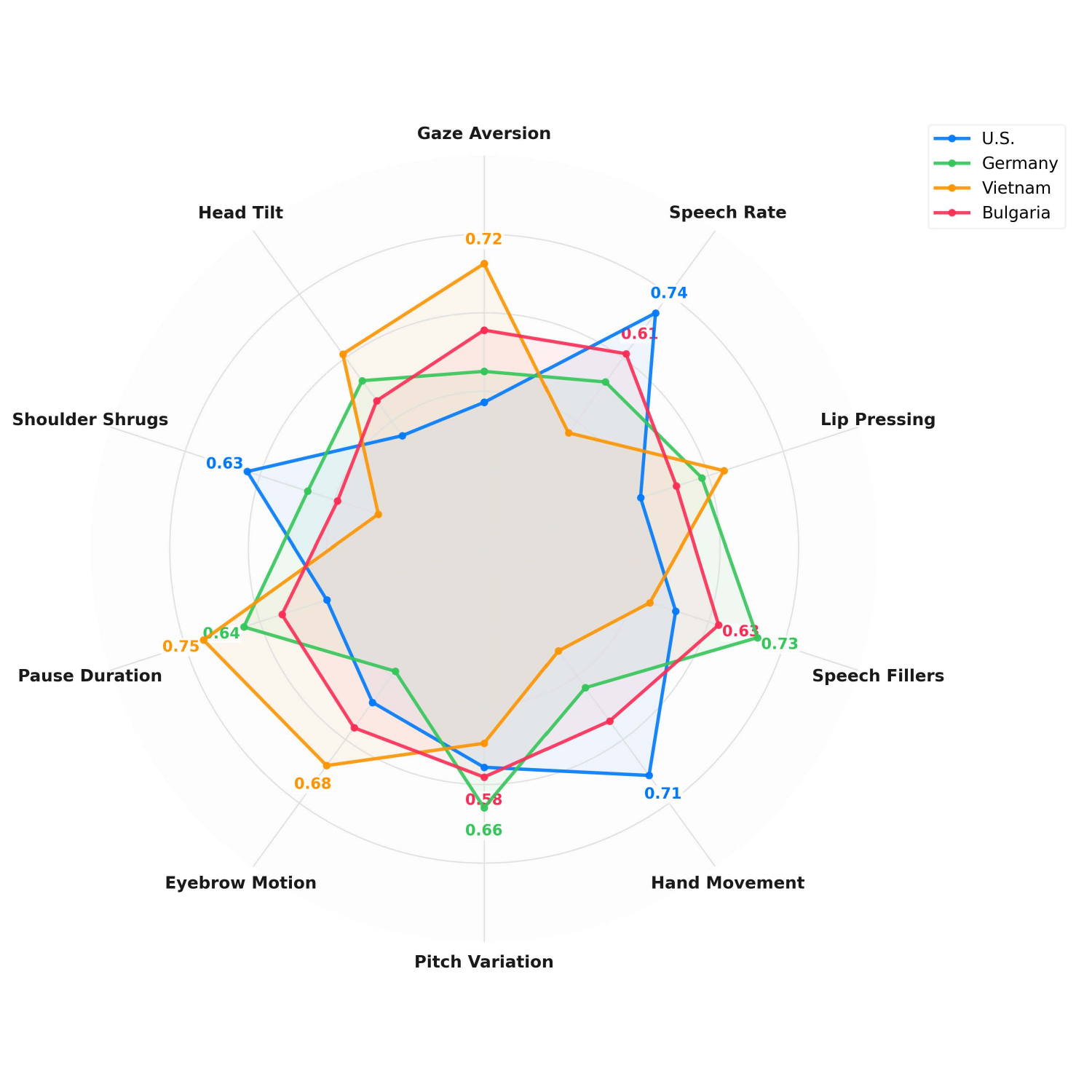}
    \end{minipage}
    \hfill
    \begin{minipage}[c]{0.35\linewidth}
        \caption{Cross-cultural distribution of deceptive signatures across 10 behavioral dimensions.}
        \label{fig:radar_chart}
    \end{minipage}
\end{figure}

This appendix provides additional details on the construction and analysis of the \textbf{T4-Deception} dataset ($N=1,695$) in four aspects:

\textbf{Source Collection and Filtering}: We describe the public sources, clip-level filtering criteria, and label verification procedure used to construct T4-Deception (Section~\ref{app:t4_filtering}).

\textbf{Cross-Cultural Behavioral Dynamics}: We analyze cross-cultural differences in reasoning cue utilization and examine how these patterns relate to dynamic modality reliance in our model (Section~\ref{subsec:cultural_cues}).

\textbf{Baseline Evaluation}: We evaluate classic deception detection methods on T4-Deception, establishing task-specific baselines for future research (Section~\ref{app:benchmark}).

\vspace{-4pt}
\subsection{Source Collection and Filtering}
\label{app:t4_filtering}
\vspace{-4pt}

T4-Deception is collected from publicly accessible online sources under the unified \textit{To Tell the Truth} format. To ensure that each clip contains clear and attributable audiovisual evidence, we retain only clips satisfying the following criteria: (i) the target subject's upper body is visible; (ii) the audio is audible and contains no other speaker; (iii) only the target subject appears in the frame; and (iv) the truthful/deceptive label is revealed by the original show. All retained clips are cross-checked by annotators, and clips shorter than 0.9s are discarded. These filtering rules reduce ambiguity in cue extraction and ensure that each sample can be associated with a single target subject and a verified ground-truth label.

\vspace{-4pt}
\subsection{Cross-Cultural Behavioral Dynamics}
\label{subsec:cultural_cues}
\vspace{-4pt}
We quantify deceptive cues through ten visual and acoustic features: Gaze Aversion, Speech Rate, Lip Pressing, Speech Fillers, Hand Movement, Pitch Variation, Eyebrow Motion, Pause Duration, Shoulder Shrugs, Head Tilt.

\textbf{Cue Distribution}
Figure~\ref{fig:radar_chart} visualizes the different cultural behavioral patterns: the U.S. subjects exhibit higher \textit{Hand Movement} intensity, whereas the Vietnamese subjects show pronounced \textit{Pause Duration}.

\begin{figure}[t]
    \centering
    \begin{minipage}[c]{0.6\linewidth}
        \centering
        \includegraphics[width=\linewidth]{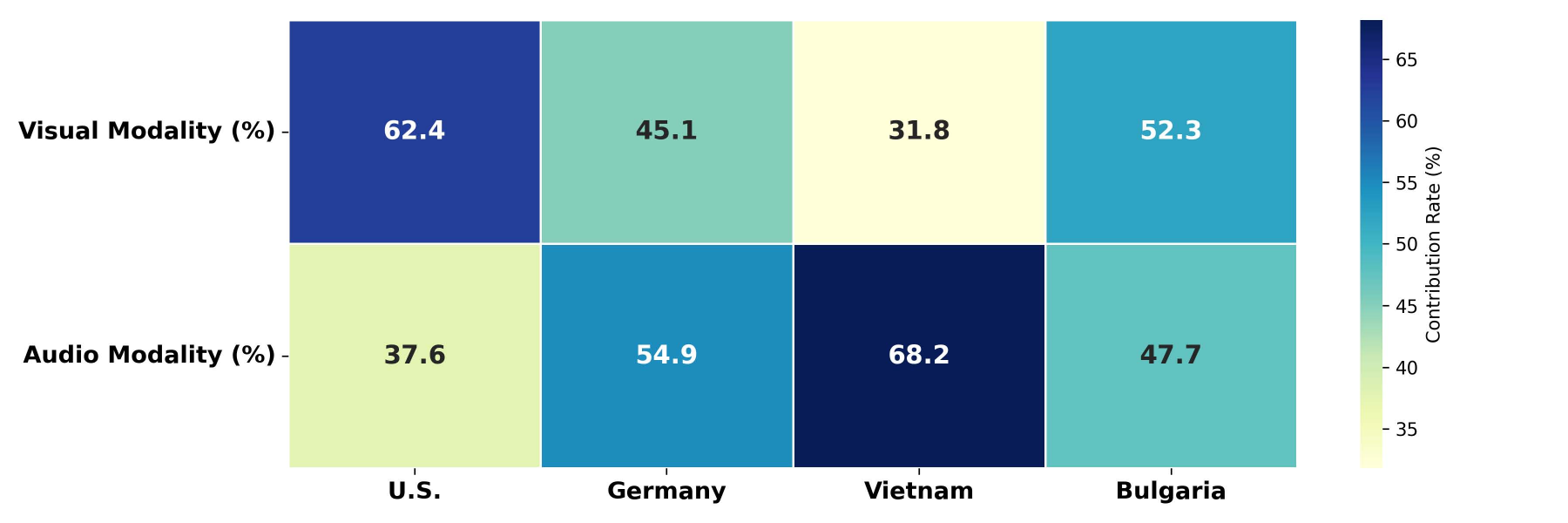}
    \end{minipage}
    \hfill
    \begin{minipage}[c]{0.35\linewidth}
        \caption{Cross-cultural modality contribution rates. The model dynamically shifts reliance between visual and audio streams.}
        \label{fig:modality_heatmap_transposed}
    \end{minipage}
\end{figure}

\textbf{Dynamic Modality Reliance}
Since deception cues vary by culture, DecepGPT automatically adjusts which signals it trusts most. As shown in Figure~\ref{fig:modality_heatmap_transposed}, the model relies primarily on \textbf{video} (62.4\%) for \textbf{U.S.} subjects. In contrast, for \textbf{Vietnamese} subjects, it shifts focus to \textbf{audio} (68.2\%), where speech pauses serve as key indicators. Meanwhile, for subjects from \textbf{Germany and Bulgaria}, the model adopts a \textbf{balanced mix} of both video and audio cues.

\vspace{-4pt}
\subsection{Baseline Evaluation}
\label{app:benchmark}
\vspace{-4pt}
We evaluate DecepGPT against task-specific baselines as shown in Table~\ref{tab:multicultural_dl_comp}.

\begin{table}[t]
\centering
\caption{Comparison with task-specific baselines in T4-deception on accuracy.}
\label{tab:multicultural_dl_comp}
\small
\begin{tabular}{l ccccc}
\toprule
\textbf{Method} & \textbf{U.S.} & \textbf{Germany} & \textbf{Vietnam} & \textbf{Bulgaria} & \textbf{Avg.} \\
\midrule
LieNet      & 52.41 & 50.85 & 48.23 & 51.16 & 50.66 \\
FacialCueNet & 54.12 & 52.60 & 49.54 & 53.08 & 52.3 \\
PECL             & 56.45 & 54.12 & 51.76 & 55.42 & 54.44 \\
\midrule
\rowcolor{gray!10} \textbf{DecepGPT} & \textbf{64.65} & \textbf{61.11} & \textbf{63.10} & \textbf{61.46} & \textbf{62.58} \\
\bottomrule
\end{tabular}
\end{table}

\vspace{-4pt}
\section{Broader Evaluation on Real-Life Trials}
\label{app:rlt_evaluation}
\vspace{-4pt}

To further evaluate the generalization ability of DecepGPT beyond laboratory and game-show scenarios, we additionally conduct experiments on the Real-Life Trials (RLT) dataset, which contains real courtroom deception cases. As shown in Table~\ref{tab:rlt_results}, DecepGPT achieves the best performance among all compared methods, reaching 87.82\% accuracy and 88.54\% F1. These results suggest that our framework remains effective in real-world courtroom scenarios, where audiovisual deception cues are more complex and less controlled than in laboratory settings.

\begin{table}[t]
\centering
\begin{minipage}[c]{0.32\linewidth}
    \caption{Evaluation on the Real-Life Trials (RLT) dataset.}
    \label{tab:rlt_results}
\end{minipage}
\hfill
\begin{minipage}[c]{0.62\linewidth}
\centering
\small
\setlength{\tabcolsep}{6pt}
\renewcommand{\arraystretch}{1.08}
\begin{tabular}{lcc}
\toprule
\textbf{Method} & \textbf{Acc.} & \textbf{F1} \\
\midrule
LieNet & 67.33 & 69.84 \\
PECL   & 71.00 & 71.21 \\
AFFAKT & 86.70 & 87.60 \\
Ours   & \textbf{87.82} & \textbf{88.54} \\
\bottomrule
\end{tabular}
\end{minipage}
\end{table}

\vspace{-4pt}
\section{Cross-Domain Ablation Study of SICS and DMC}
\label{app:cross_domain_ablation}
\vspace{-4pt}
Table~\ref{tab:cross_domain_ablation_results} details how SICS or DMC affects the model's ability to generalize across different datasets, while the main text focuses on in-domain evaluation. 

\begin{table}[t] 
\centering
\caption{Cross-domain ablation study on core modules. ``X\&Y$\rightarrow$Z'' denotes training on datasets X and Y, testing on the held-out target Z.}
\label{tab:cross_domain_ablation_results}
\renewcommand{\arraystretch}{1.5} 
\footnotesize 
\setlength{\tabcolsep}{3pt} 

\begin{tabular}{l cccccccc}
\toprule
\multirow{2}{*}{\textbf{Variant}} & \multicolumn{2}{c}{M\&B $\rightarrow$ D} & \multicolumn{2}{c}{D\&M $\rightarrow$ B} & \multicolumn{2}{c}{D\&B $\rightarrow$ M} & \multicolumn{2}{c}{\textbf{Average}} \\ 
\cmidrule(lr){2-3} \cmidrule(lr){4-5} \cmidrule(lr){6-7} \cmidrule(lr){8-9}
 & Acc. & F1 & Acc. & F1 & Acc. & F1 & Acc. & F1 \\ 
\midrule
Base (No SICS/DMC) & 54.12 & 58.20 & 50.50 & 64.50 & 51.20 & 54.30 & 51.94 & 59.00 \\
Base + DMC         & 56.45 & 59.88 & 53.21 & 66.75 & 52.84 & 56.12 & 54.17 & 60.92 \\
Base + SICS        & 61.08 & 60.43 & 57.44 & 69.82 & 55.60 & 59.15 & 58.04 & 62.13 \\
\midrule
\rowcolor{gray!10} \textbf{Full Model} & \textbf{63.46} & \textbf{61.79} & \textbf{59.62} & \textbf{72.00} & \textbf{57.24} & \textbf{60.38} & \textbf{60.11} & \textbf{64.72} \\ 
\bottomrule
\end{tabular}
\end{table}

\vspace{-4pt}
\section{Detailed Component Analysis and Backbone Comparison}
\label{app:component_analysis}
\vspace{-4pt}
We provide the comprehensive component analysis and backbone comparison results for the Bag-of-Lies and MU3D datasets as shown in Table \ref{app:sics_components} and Table \ref{app:backbone_comp}, complementing the DOLOs results shown in the main text.

\vspace{-4pt}
\subsection{Fine-Grained Ablation of SICS Components}
\label{app:sics_components}
\vspace{-4pt}
Table~\ref{tab:extended_sics_ablation} confirms that removing any SICS component (global prior, polarity-aware adjustment, or gating) consistently degrades performance across both datasets, validating the necessity of each module.

\begin{table}[t]
    \centering
    \begin{minipage}[c]{0.25\textwidth} 
        \caption{Ablation study on SICS internal components across standard benchmarks.}
        \label{tab:extended_sics_ablation}
    \end{minipage}
    \hfill
    \begin{minipage}[c]{0.72\textwidth}
        \centering
        \small
        \renewcommand{\arraystretch}{1.6} 
        \setlength{\tabcolsep}{4.5pt} 
        
        \begin{tabular}{l cc cc cc}
            \toprule
            & \multicolumn{2}{c}{\textbf{DOLOs}} & \multicolumn{2}{c}{\textbf{Bag-of-Lies}} & \multicolumn{2}{c}{\textbf{MU3D}} \\
            \cmidrule(lr){2-3} \cmidrule(lr){4-5} \cmidrule(lr){6-7} 
            \textbf{Variant} & \textbf{Acc.} & \textbf{F1} & \textbf{Acc.} & \textbf{F1} & \textbf{Acc.} & \textbf{F1}\\
            \midrule
            w/o $\mathbf{b}_{global}$  & 71.42 & 74.31 & 61.22 & 61.05 & 58.85 & 64.12 \\
            w/o Polarity      & 72.06 & 75.28 & 60.20 & 59.40 & 59.40 & 65.25 \\
            w/o Gating        & 71.85 & 74.94 & 61.85 & 61.32 & 59.10 & 65.50 \\
            \midrule
            \rowcolor{gray!10} \textbf{Full SICS} & \textbf{73.23} & \textbf{76.13} & \textbf{63.46} & \textbf{63.72} & \textbf{61.25} & \textbf{67.22} \\
            \bottomrule
        \end{tabular}
    \end{minipage}
\end{table}

\begin{table}[t]
    \centering
    \begin{minipage}[c]{0.25\textwidth} 
        \caption{Comparison of backbone architectures on standard benchmarks. The LLM-based approach significantly outperforms traditional deep learning backbones.}
        \label{tab:extended_backbone}
    \end{minipage}
    \hfill
    \begin{minipage}[c]{0.72\textwidth}
        \centering
        \small
        \renewcommand{\arraystretch}{1.6} 
        \setlength{\tabcolsep}{4.5pt} 
        
        \begin{tabular}{l cc cc cc}
            \toprule
            & \multicolumn{2}{c}{\textbf{DOLOs}} & \multicolumn{2}{c}{\textbf{Bag-of-Lies}} & \multicolumn{2}{c}{\textbf{MU3D}} \\
            \cmidrule(lr){2-3} \cmidrule(lr){4-5} \cmidrule(lr){6-7} 
            \textbf{Backbone} & \textbf{Acc.} & \textbf{F1} & \textbf{Acc.} & \textbf{F1} & \textbf{Acc.} & \textbf{F1}\\
            \midrule
            MLP         & 58.42 & 60.17 & 51.25 & 48.90 & 52.10 & 56.45 \\
            Transformer & 64.75 & 68.34 & 56.84 & 55.42 & 55.62 & 61.08 \\
            \midrule
            \rowcolor{gray!10} \textbf{DecepGPT} & \textbf{73.23} & \textbf{76.13} & \textbf{63.46} & \textbf{63.72} & \textbf{61.25} & \textbf{67.22} \\
            \bottomrule
        \end{tabular}
    \end{minipage}
\end{table}

\begin{figure}[!h]
  \centering
  \includegraphics[width=1\columnwidth]{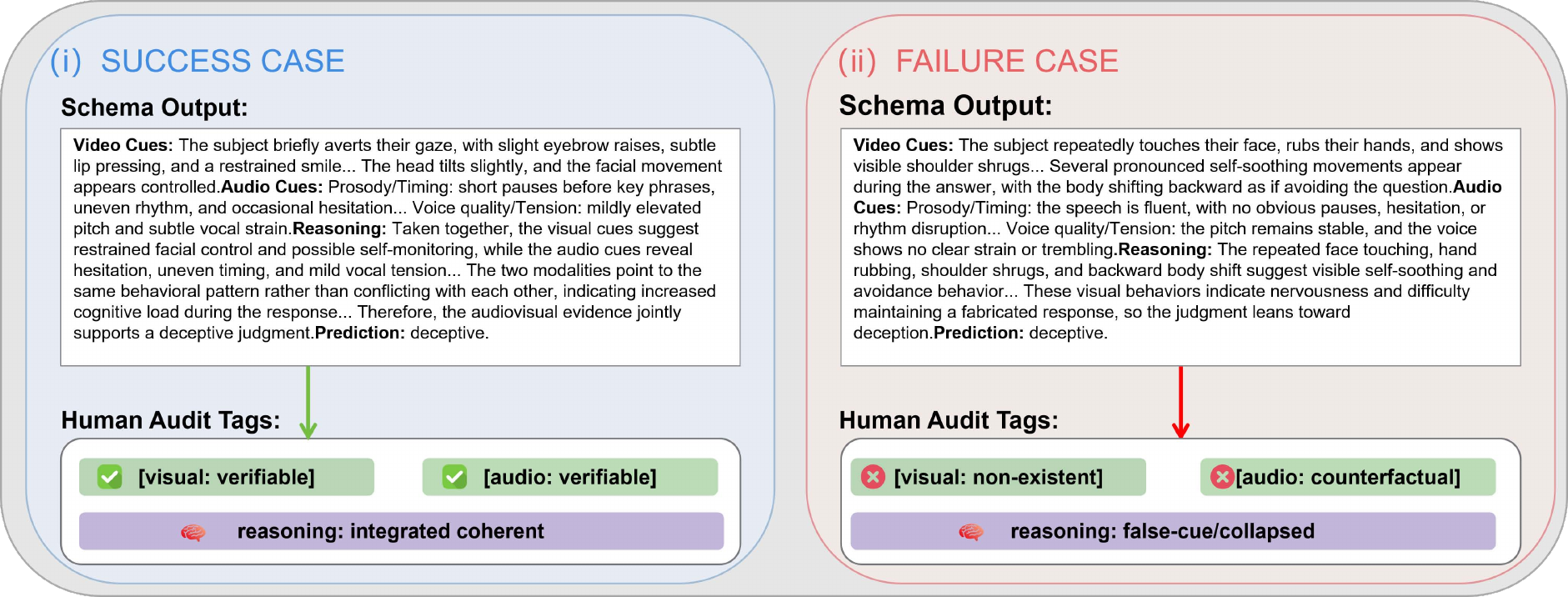} 
    \caption{Abridged qualitative comparison of a success case and a failure case with human audit tags.}
  \label{fig:qualitative_sample}
\end{figure}

\vspace{-4pt}
\subsection{Backbone Architecture Comparison}
\label{app:backbone_comp}
\vspace{-4pt}
Table~\ref{tab:extended_backbone} shows that the LLM backbone significantly outperforms MLP and Transformer alternatives on Bag-of-Lies and MU3D, highlighting the critical role of advanced semantic reasoning in multimodal deception detection.

\vspace{-4pt}
\section{Qualitative Analysis of Auditable Reasoning}
\label{app:qualitative_analysis}
\vspace{-4pt}
To complement the quantitative audit results in the main text, we provide a qualitative case study of the generated audit reports. For readability and space constraints, Fig.~\ref{fig:qualitative_sample} shows abridged excerpts rather than complete model outputs, while preserving the key cues, reasoning steps, predictions, and human audit tags. The success case illustrates how the model connects observed audiovisual cues to a coherent prediction, whereas the failure case shows how hallucinated or counterfactual cues can lead to an unreliable rationale. This comparison illustrates the value of schema-constrained reasoning: the generated reports expose the evidentiary basis of each prediction, making both correct decisions and failure modes easier to inspect. These artifacts support post-hoc human verification of model logic without requiring access to internal black-box states.

\vspace{-4pt}
\section{Role of GPT-4o in This Work}
\label{app:gpt4o_role}
\vspace{-4pt}

GPT-4o is used in this work from three complementary perspectives: data annotation, report evaluation, and commercial-model comparison.

First, GPT-4o assists the construction of high-quality supervision signals during data annotation. This stage is not performed as a single end-to-end generation process. Instead, GPT-4o is used in several independent steps with different functional roles. Specifically, GPT-4o is first used to generate structured visual cue descriptions from video frames, including gaze behavior, facial movement, head motion, posture, and other observable nonverbal signals. Based on the generated audiovisual cue descriptions, GPT-4o is then used to generate reasoning chains. Finally, GPT-4o is used for semantic-preserving rewriting to produce augmented report samples with diverse expressions, while preserving the semantics of the cues and reasoning. Human annotators further review these independently generated outputs to ensure the accuracy of the cues and reasoning chains.

Second, GPT-4o is used as an automatic evaluator for model-generated audit reports. Inspired by recent LLM-as-a-judge evaluation protocols, we use GPT-4o to intelligently compare the model-generated reports with the ground-truth reports. Specifically, GPT-4o evaluates whether the generated report is semantically aligned with the ground truth in terms of behavioral cue descriptions and reasoning-chain. 

Third, GPT-4o serves as a strong commercial closed-source baseline for end-to-end deception detection. Since GPT-4o is one of the most capable general-purpose multimodal models, comparing it with DecepGPT allows us to examine whether a general MLLM can directly solve multimodal deception detection. This comparison also helps reveal whether the proposed task-specific framework provides additional benefits beyond general multimodal understanding. The results highlight the gap between broad multimodal reasoning capability and robust, evidence-traceable deception judgment in specialized scenarios.

\end{document}